\documentclass[letterpaper]{article}
\usepackage[preprint]{aaai2027}

\usepackage[hyphens]{url}
\usepackage{graphicx}
\def\UrlFont{\rm}

\usepackage{natbib}
\usepackage{caption}
\usepackage{subcaption}
\usepackage{booktabs}
\usepackage{amsmath}
\usepackage{amssymb}
\usepackage{comment}
\usepackage{bm}
\usepackage{mathrsfs}
\usepackage[cal=cm]{mathalpha}
\usepackage{makecell}
\usepackage{todonotes}
\usepackage{enumitem}
\usepackage{colortbl}
\usepackage{xcolor}
\usepackage{siunitx}
\usepackage{multirow}

\definecolor{lightgray}{gray}{0.85}
\newcommand{\vect}[1]{\mathbf{#1}}
\newcommand{\R}{\mathbb{R}}
\newcommand{\K}{\mathord{\scalebox{0.8}{$\mathcal{K}$}}}

\newcommand{\PaperTitle}{%
From Digital to Physical Reservoir Computing:\\
Co-Optimizing Soft Robotic Reservoirs via Dynamics Matching%
}

\author{%
  Nicola Visentin\corresponding\textsuperscript{\rm 1,\rm 3},
  Maximilian Stölzle\textsuperscript{\rm 1,\rm 2},
  Mariano Ramírez Montero\textsuperscript{\rm 1},\\
  Francesco Braghin\textsuperscript{\rm 3},
  Daniela Rus\textsuperscript{\rm 2},
  Cosimo Della Santina\textsuperscript{\rm 1}
}
\affiliations{%
  \textsuperscript{\rm 1}Department of Cognitive Robotics,
  Delft University of Technology, Delft, The Netherlands\\
  \textsuperscript{\rm 2}Computer Science and Artificial Intelligence Laboratory,
  Massachusetts Institute of Technology, Cambridge, MA, USA\\
  \textsuperscript{\rm 3}Department of Mechanical Engineering,
  Politecnico di Milano, Milan, Italy\\
  nicola.visentin@mail.polimi.it,
  mstolzle@mit.edu,
  m.ramirezmontero-1@tudelft.nl\\
  francesco.braghin@polimi.it,
  rus@csail.mit.edu,
  c.dellasantina@tudelft.nl
}

\title{\PaperTitle}

\begin{document}

\maketitle

\begin{abstract}
Soft robotic substrates are promising for Physical Reservoir Computing (PRC) because their compliant nonlinear dynamics can provide temporal memory, high-dimensional state transformations, and efficient inference. However, physical reservoirs are often adopted as-is rather than pretrained or co-optimized, potentially limiting soft robotic PRC performance relative to digital reservoirs. We investigate whether a physical reservoir can instead be pretrained against high-performing digital reference dynamics. Our formulation jointly optimizes physical parameters, a diffeomorphic physical--reference state map, and feedforward-feedback control using a differentiable physical model and an acceleration-level equation-error objective that avoids temporal integration. As a proof of concept, we instantiate the formulation with simulated soft robots, a Random Oscillators Network (RON) reference, and parallel multi-start gradient descent. We evaluate the optimized reservoirs on classification (sMNIST and ADIAC) and forecasting (Mackey--Glass and Lorenz96) tasks across four reservoir dimensions. Compared with unoptimized soft robot reservoirs, the optimized reservoirs achieve a mean relative improvement of \SI{33.7}{\percent} across all tasks and datasets, while remaining close to the digital reference. These results demonstrate the feasibility of dynamics-level co-optimization for the simulated soft robotic reservoirs considered here.
\end{abstract}

\section{Introduction}
\label{sec:introduction}

Reservoir Computing (RC)~\cite{RC_book,reservoir_computing_approaches_to_recurrent_nn_training} processes temporal data with a fixed nonlinear dynamical system, the \textit{reservoir}, followed by a trainable readout layer. The reservoir expands the input history into a rich state representation, while learning is confined to the readout. As a result, RC avoids backpropagation through the full recurrent dynamics and can be substantially cheaper to train than conventional Recurrent Neural Networks (RNNs)~\cite{BPTT}. Physical Reservoir Computing (PRC) extends this idea by replacing the simulated reservoir with a physical substrate~\cite{physical_reservoir_computing_an_introductory_prospective,RC_book:reservoir_computing_in_material_substrate}. Since the reservoir dynamics unfold directly in the physical medium rather than being numerically simulated, PRC can reduce digital computation during inference and is therefore attractive for efficient real-time or embodied processing. Among the many substrates explored for PRC~\cite{recent_advances_in_prc_a_review}, soft robots are especially promising because their compliant, coupled, nonlinear dynamics can naturally provide temporal memory and high-dimensional state transformations~\cite{RC_book:physical_reservoir_computing_in_robotics,towards_a_theoretical_foundation_for_morphological_computation_with_compliant_bodies}.

Despite this promise, current PRC systems still lag behind digital reservoirs. In soft robotic PRC, existing demonstrations typically address relatively simple tasks and often use the physical body as-is, whereas digital reservoirs are explicitly selected or tuned for RC~\cite{information_processing_via_physical_soft_body,PRC_in_a_soft_swimming_robot}. Existing reservoir-improvement methods offer limited guidance for this setting: many are tailored to specific digital architectures, rely on hand-tuned heuristics, or require repeated downstream task evaluations~\cite{reservoir_computing_approaches_to_recurrent_nn_training}. Likewise, generic soft robot co-design methods~\cite{multi_objective_design_of_a_soft_pneumatic_robot,soft_robot_holistic_co_design,curriculum_based_co_design_of_morphology_etc} do not directly resolve the PRC design problem. Optimizing the morphology and control for final task performance would require rolling out each candidate physical reservoir, training the readout, and differentiating through the integrated dynamics of the full PRC pipeline. Replacing this with proxy or surrogate objectives such as expressiveness or memory capacity avoids full task optimization, but makes the design criterion indirect and partly arbitrary. This leaves a clear gap: PRC lacks a tractable way to design physical reservoirs whose dynamics are useful for RC without optimizing through every downstream task.

\begin{figure}[!t]
    \centering
    \includegraphics[width=1.0\columnwidth]{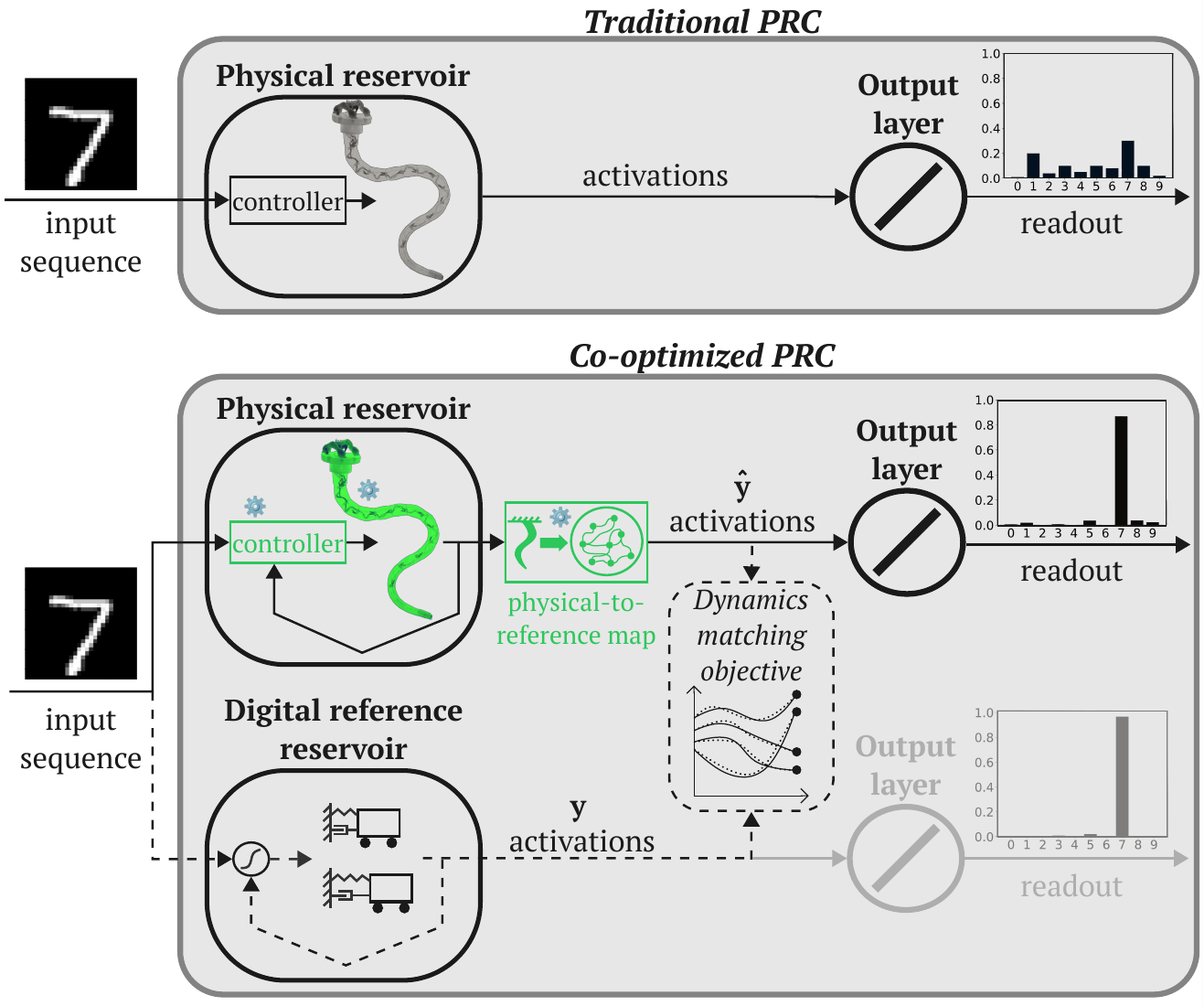}
    \caption{Traditional PRC (top) and the proposed pretrained physical reservoir (bottom), illustrated for a soft robot handwritten-digit recognition task. In traditional PRC, the input sequence drives an unoptimized reservoir whose measured outputs form the readout activations. In the proposed approach, the physical parameters, controller, and map are co-optimized such that the mapped physical dynamics match those of a high-performing digital reference reservoir. Solid and dashed lines denote inference and pretraining, respectively; green elements and the gear symbol indicate pretrained components.}
    \label{fig:methodology_high_level}
\end{figure}

We address this gap by investigating whether a physical reservoir can be pretrained to approximate a high-performing digital reference reservoir. For differentiable second-order systems, we jointly adapt the physical parameters, a diffeomorphic physical--reference state map, and feedforward-feedback control using an acceleration-level equation-error objective~\cite{robot_model_identification_and_learning_ecc}. A differentiable physical model provides gradients without temporal integration. As a proof of concept, we use a physics-inspired Random Oscillators Network (RON)~\cite{random_oscillators_network_for_time_series_processing} as reference and simulated Lagrangian soft robots as physical reservoirs, optimized with parallel multi-start gradient descent. After pretraining, the optimized reservoir is inserted into a standard PRC pipeline, and only the readout is trained. The full pipeline remains hybrid: its controller, map, and readout are digital. Fig.~\ref{fig:methodology_high_level} provides an overview.

We evaluate this proof of concept in simulation. The contributions are:
\begin{enumerate}
    \item We introduce a dynamics-level pretraining perspective in which a physical reservoir is co-optimized against structurally compatible, high-performing digital reference dynamics rather than treated as fixed or selected as-is.
    \item For differentiable second-order systems, we formulate this idea using a digital reference reservoir, a diffeomorphic physical--reference state map, feedforward-feedback controllers co-optimized with physical parameters, and an acceleration-level dynamics-matching objective that avoids temporal rollouts.
    \item We instantiate and evaluate the formulation for simulated soft robotic reservoirs using parallel multi-start gradient descent, providing a proof of concept on classification and forecasting tasks.
\end{enumerate}

\section{Related Work}
\label{sec:related_work}

\subsubsection{Physical Reservoir Computing} PRC exploits the fact that the reservoir is a fixed nonlinear dynamical system, and thus any physical medium with rich, history-dependent dynamics can replace a software-simulated reservoir~\cite{physical_reservoir_computing_an_introductory_prospective,RC_book:reservoir_computing_in_material_substrate}. This follows the principles of morphological computation~\cite{morphological_computation_the_body_as_a_computational_resource,morphological_computation_and_morphological_control}, where the body performs part of the computation and reduces digital processing.

Many substrates have been explored for PRC, including photonic systems~\cite{PRC_photonics}, nanomaterials~\cite{PRC_nanomaterials}, MEMS~\cite{PRC_mems}, biological tissue~\cite{PRC_biology}, quantum dynamics~\cite{PRC_quantum_dynamics} and others surveyed in~\cite{recent_advances_in_prc_a_review}. Soft robots are especially relevant~\cite{RC_book:physical_reservoir_computing_in_robotics}, since their compliant, coupled nonlinear dynamics can naturally provide the high-dimensional state representations and memory required for RC~\cite{towards_a_theoretical_foundation_for_morphological_computation_with_compliant_bodies}.

Real-world soft robot PRC has been demonstrated, for example in~\cite{information_processing_via_physical_soft_body,PRC_in_a_soft_swimming_robot}. Nevertheless, these systems mostly solve simple tasks and still underperform compared to digital reservoirs, which are explicitly tailored for RC. Since designing a physical substrate for reservoir performance is harder than tuning a digital reservoir, the physical reservoir is usually employed as-is, without pretraining or pre-selection.

\subsubsection{Reservoir Pretraining} In digital RC, reservoir design usually starts from random architectures with hand-tuned global parameters~\cite{reservoir_computing_approaches_to_recurrent_nn_training}. For the well-known Echo State Networks~\cite{original_echo_state_network_article}, this includes scaling the spectral radius to promote the echo state property~\cite{echo_state_network_tutorial}; more generally, topology, timescales, and architecture-specific parameters can be tuned by search~\cite{exploration_of_effects_of_different_network_topologies_on_esn,optimization_of_leaky_esn,esn_with_band_pass_neurons,random_oscillators_network_for_time_series_processing}. Other approaches optimize task-independent proxy objectives that capture desirable reservoir properties, such as memory, separation, activation entropy, or self-prediction~\cite{improving_reservoirs_using_intrinsic_plasticity,esn_and_self_prediction}. Finally, task-driven adaptation uses labels or downstream performance to tune weights, topology, or global parameters, typically through an outer optimization loop with repeated task evaluations~\cite{supervised_and_evolutionary_learning_of_esn,identification_of_motion_with_esn}.

These methods offer limited guidance for PRC because they are architecture-specific, proxy-based, or require repeated task evaluations. Viewed as teacher-student dynamics distillation~\cite{hinton2015distilling}, our method transfers a digital reservoir's vector field into a physical one without repeated task-level optimization. Related work trains recurrent networks from teacher-generated dynamics (full-FORCE)~\cite{depasquale2018full_force}, or mirrors a digital ESN in analog spiking hardware~\cite{he2019reservoir_transfer}; we instead co-optimize physical morphology, a state map, and control through offline equation error. Unlike MRAC~\cite{landau1979adaptive_control}, this requires neither online adaptation nor the reference at inference.

\section{Physical Reservoir Pretraining}
\label{sec:methodology}

\begin{figure*}[!t]
    \centering
    \includegraphics[width=\textwidth]{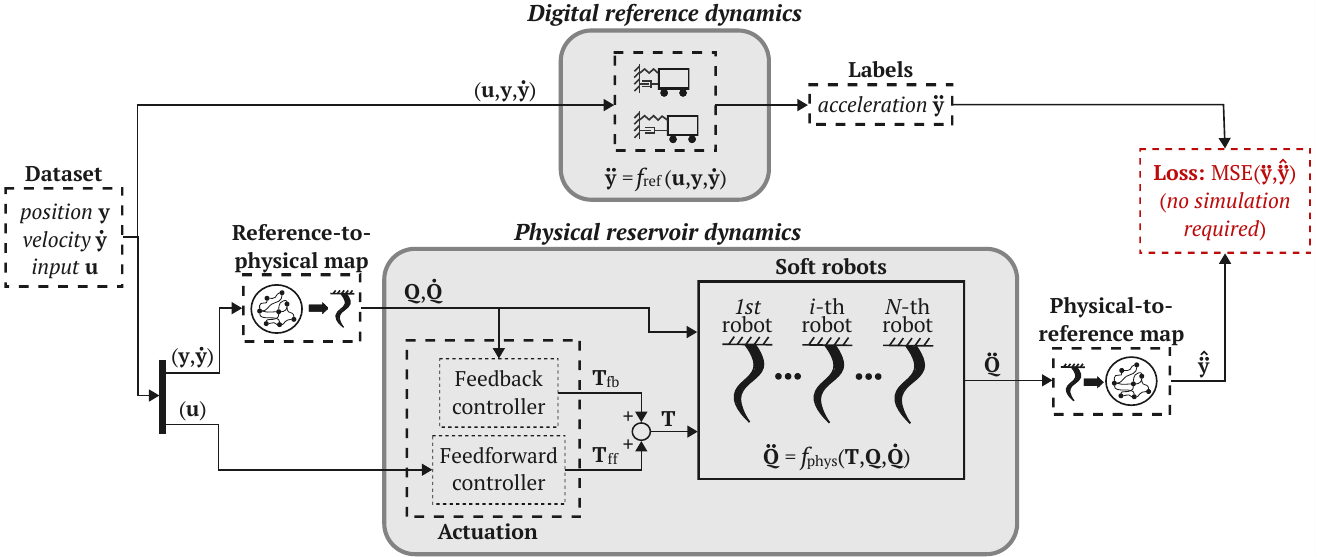}
    \caption{Soft robot instantiation of the proposed pretraining formulation. The reference position $\vect{y}$, velocity $\vect{\dot{y}}$, and input $\vect{u}$ are mapped into the physical space; the resulting acceleration $\vect{\ddot{Q}}$ is mapped back to the reference space and compared with the target acceleration $\vect{\ddot{y}}$. Dashed lines denote digital components, while solid lines indicate elements to be implemented physically.}
    \label{fig:methodology}
\end{figure*}

This section formulates the pretraining approach for differentiable second-order physical reservoirs with coordinate spaces compatible with a diffeomorphism. A digital reference reservoir supplies target dynamics, while the physical parameters, state map, and controller are jointly adapted. Sec.~\ref{sec:optimization_experiments} presents one simulated soft robot instantiation; Fig.~\ref{fig:methodology} illustrates its pretraining setup.

\subsection{Digital Reference Reservoir}
\label{subsec:reference_reservoir}

The formulation requires a digital reservoir whose dynamics define the optimization target. The reference should exhibit strong task performance and structural compatibility with the physical substrate, enabling transfer of task-relevant dynamics.

We consider reference dynamics of the second-order form
\begin{equation}
    \vect{\ddot{y}} = f_\mathrm{ref}(\vect{y}, \vect{\dot{y}}, \vect{u}),
    \label{eq:reference_dynamics_generic}
\end{equation}
where $\vect{y}(t)\in\R^{n_y}$ are the reference coordinates, $(\vect{y},\vect{\dot y})$ is the reference state, and $\vect{u}(t)\in\R^{n_u}$ is the input. The specific instantiation of $f_\mathrm{ref}$ used here is discussed in Sec.~\ref{subsec:optimiz_setup}.

\subsection{Differentiable Physical Reservoir}
\label{subsec:physical_reservoir}

Let $\vect{Q}(t)\in\R^{n_Q}$ denote aggregate physical reservoir coordinates, $\vect{T}(t)\in\R^{n_T}$ the actuation, and $\theta_\mathrm{phys}$ trainable physical parameters. We assume a differentiable second-order model
\begin{equation}
    \vect{\ddot Q}=f_\mathrm{phys}(\vect{Q},\vect{\dot Q},\vect{T};\theta_\mathrm{phys}).
    \label{eq:physical_dynamics_generic}
\end{equation}
Define the state $\vect{Z}=[\vect{Q}^\top,\vect{\dot Q}^\top]^\top\in\R^{2n_Q}$.

\subsubsection{Controller} The input $\vect{u}(t)\in\R^{n_u}$ reaches the physical reservoir through a feedforward term $\vect{T}_\mathrm{ff}(\vect{u})$. A feedback term $\vect{T}_\mathrm{fb}(\vect{Z})$ introduces dynamical coupling and augments reservoir expressiveness. The complete actuation $\vect{T}(t)$ is
\begin{equation}
    \vect{T}(\vect{Z},\vect{u}; \theta_\mathrm{contr})=\vect{T}_\mathrm{fb}(\vect{Z}; \theta_\mathrm{contr}^{(\mathrm{fb})})+\vect{T}_\mathrm{ff}(\vect{u}; \theta_\mathrm{contr}^{(\mathrm{ff})}),
    \label{eq:controller_law_generic}
\end{equation}
where $\theta_\mathrm{contr}=\{ \theta_\mathrm{contr}^{(\mathrm{fb})}, \theta_\mathrm{contr}^{(\mathrm{ff})} \}$ are learnable parameters.

\subsection{Physical--Reference Reservoir Mapping}
\label{subsec:mapping}

The digital reference and physical coordinates may differ in meaning and scale. We therefore require $n_y=n_Q$ and introduce a map $\phi:\R^{n_y}\rightarrow \R^{n_Q}$:
\begin{equation}\label{eq:y2q_and_q2y}
    \vect{Q}=\phi(\vect{y}; \theta_\mathrm{map}),
    \quad
    \vect{y}=\phi^{-1}(\vect{Q}; \theta_\mathrm{map})=\psi(\vect{Q}; \theta_\mathrm{map}).
\end{equation}
$\theta_\mathrm{map}$ are learnable parameters. The map must be an invertible, second-order smooth diffeomorphism so positions, velocities, and accelerations can be transformed consistently through the Jacobians and Hessians of Eq.~\eqref{eq:y2q_and_q2y}.

More expressive maps can improve dynamics matching, while simpler maps reduce digital computation. Sec.~\ref{subsec:optimiz_setup} specifies the map used in our experiments; further details are provided in the supplementary material.

\subsection{Optimization Strategy}
\label{subsec:optimization_strategy}

Let $\vect{y}$ denote the digital reference coordinates and $\vect{\hat y}=\phi^{-1}(\vect{Q})$ the mapped physical coordinates. Pretraining adapts $\theta=\{ \theta_\mathrm{phys},\theta_\mathrm{map},\theta_\mathrm{contr} \}$, namely the physical, map, and controller parameters. The equation-error loss~\cite{robot_model_identification_and_learning_ecc} compares accelerations and avoids temporal integration:
\begin{equation}
    \begin{array}{ll}
    \displaystyle \min_{\theta} & \mathcal{L}(\theta), \\[6pt]
    \text{s.t.}                   & \theta_\mathrm{phys}\in\mathcal{C}_\mathrm{phys}, \quad \theta_\mathrm{map}\in\mathcal{C}_\mathrm{map}, \\[4pt]
    \end{array}
    \label{eq:optimization_problem}
\end{equation}
with loss function $\mathcal{L}$ as the Mean Square Error (MSE)
\begin{equation*}
    \mathcal{L}(\theta)=\mathrm{MSE}(\vect{\ddot{y}}, \vect{\hat{\ddot{y}}}(\theta))=\frac{1}{n_\mathrm{batch}}\sum_{i=1}^{n_\mathrm{batch}} \left\| \vect{\ddot{y}}^{(i)}- \vect{\hat{\ddot{y}}}^{(i)}(\theta) \right\|^2.
\end{equation*}
For samples $(\vect{y},\vect{\dot y},\vect{u})$, labels $\vect{\ddot y}$ follow from Eq.~\eqref{eq:reference_dynamics_generic}, while the mapped physical acceleration $\vect{\hat{\ddot y}}$---obtained from Eqs.~\eqref{eq:physical_dynamics_generic}, \eqref{eq:controller_law_generic} and \eqref{eq:y2q_and_q2y} with its derivatives---is
\begin{equation}
    \begin{gathered}
    \vect{\hat{\ddot Q}}
    =f_\mathrm{phys}\bigl(\vect{Q}_\phi,\vect{\dot Q}_\phi,
        \vect{T}(\vect{Z}_\phi,\vect{u})\bigr), \\
    \vect{\hat{\ddot y}}
    =\mathrm{D}\psi(\vect{Q}_\phi)[\vect{\hat{\ddot Q}}]
    + \mathrm{D}^2\psi(\vect{Q}_\phi)[\vect{\dot Q}_\phi,\vect{\dot Q}_\phi],
    \end{gathered}
    \label{eq:loss_fn}
\end{equation}
where $\vect{Q}_\phi=\phi(\vect{y})$, $\vect{\dot Q}_\phi=\mathrm{D}\phi(\vect{y})[\vect{\dot y}]$, and $\vect{Z}_\phi=[\vect{Q}_\phi^\top,\vect{\dot Q}_\phi^\top]^\top$. The expressions $\mathrm{D}f(\vect{x})[\vect{v}]$ and $\mathrm{D}^2f(\vect{x})[\vect{v},\vect{w}]$ denote the first and second Fr\'echet derivatives of $f$ at $\vect{x}$ applied to the indicated directions. The sets $\mathcal{C}_\mathrm{phys}$ and $\mathcal{C}_\mathrm{map}$ enforce physical feasibility and mapping well-posedness.

\section{Physical Reservoir Optimization Experiments}
\label{sec:optimization_experiments}

This section instantiates the pretraining formulation from Sec.~\ref{sec:methodology} with Lagrangian-inspired digital reference dynamics, soft robot model, map, controller, and optimizer used in our experiments. We then evaluate how closely the optimized physical reservoir matches the target dynamics.

\subsection{Optimization Setup} 
\label{subsec:optimiz_setup}

\subsubsection{Digital Reference Reservoir} We use the Random Oscillators Network (RON) from~\cite{random_oscillators_network_for_time_series_processing} as reference dynamics because it performs well on temporal tasks and, as a network of nonlinearly coupled oscillators, has dynamics compatible with soft robotic systems. A RON consists of $n_y$ mass-spring-damper elements coupled by a nonlinear forcing term:
\begin{equation*}
    \vect{\ddot y}=-\bm{\gamma}\odot\vect{y}-\bm{\varepsilon}\odot\vect{\dot y}+\tanh(\vect{W}\vect{y}+\vect{V}\vect{u}+\vect{b}).
\end{equation*}
Here, $\vect{y}(t)\in\R^{n_y}$ collects the oscillator positions (reservoir coordinates), $\vect{u}(t)\in\R^{n_u}$ is the input, $\bm{\gamma}\in\R^{n_y}$ and $\bm{\varepsilon}\in\R^{n_y}$ are stiffness and damping factors, and $\vect{W},\vect{V},\vect{b}$ define the nonlinear coupling and input forcing. $\odot$ denotes the element-wise product.

We consider four benchmarks involving sequential inputs or time-series prediction: sequential-MNIST (sMNIST), ADIAC, Mackey--Glass and Lorenz96. The original RON configurations in~\cite{random_oscillators_network_for_time_series_processing} use relatively large reservoirs, making differentiable soft robot optimization prohibitively expensive. We therefore use four reduced reference dimensions for each task, $n_y\in\{6,9,12,15\}$, which keeps the optimization tractable while still testing how the approach scales with reservoir size. This inevitably limits the capability of the model, but our goal is to test whether optimizing the physical reservoir can effectively improve PRC performance, rather than achieving state-of-the-art results.

For each task and reservoir dimension, the remaining RON parameters are selected through task-specific manual tuning following~\cite{random_oscillators_network_for_time_series_processing}. This yields 16 pretraining datasets, each obtained by sampling $m=10^5$ points $(\vect{y},\vect{\dot y},\vect{u})$ and computing the labels $\vect{\ddot y}$ with $f_\mathrm{ref}$. Further details on the RON model, task-specific configurations, reduced hidden space dimensionality, and dataset generation are given in the supplementary material.

\subsubsection{Soft Robot Model} Each physical reservoir element is a soft robot with generalized coordinates $\vect{q}(t)\in\R^n$ and Euler--Lagrange dynamics~\cite{della2023model}
\begin{equation*}
    \vect{M}\vect{\ddot q}+\vect{C}\vect{\dot q}+\vect{D}\vect{\dot q}+\vect{V}=\bm{\tau},
\end{equation*}
where $\vect{M}(\vect{q})$, $\vect{C}(\vect{q},\vect{\dot q})$, $\vect{D}$, and $\vect{V}(\vect{q})$ denote mass, Coriolis/centrifugal, damping, and potential terms, and $\bm{\tau}\in\R^{n}$ is the actuation. For $N$ robots, let $\vect{Q}\in\R^{Nn}$ and $\vect{T}\in\R^{Nn}$ stack their coordinates and actuation, and let $\vect{\tilde M}$, $\vect{\tilde C}$, $\vect{\tilde D}$, and $\vect{\tilde V}$ collect the corresponding dynamics. Then
\begin{equation*}
    \vect{\tilde M}\vect{\ddot Q}+\vect{\tilde C}\vect{\dot Q}
    +\vect{\tilde D}\vect{\dot Q}+\vect{\tilde V}=\vect{T},
\end{equation*}
which instantiates Eq.~\eqref{eq:physical_dynamics_generic} with $\theta_\mathrm{phys}=\theta_\mathrm{rob}$, the robot physical parameters. We model each robot as a slender continuum arm using the planar Piecewise Constant Strain (PCS) model~\cite{pcs_paper2}. Each segment contributes bending, shear, and axial strains, so a robot with $n_\mathrm{pcs}$ segments has $\vect{q}\in\R^{3n_\mathrm{pcs}}$. We choose $N$ and $n_\mathrm{pcs}$ such that $3Nn_\mathrm{pcs}=n_y$. Tab.~\ref{tab:optimization_results} reports the specific values of $N$ and $n_\mathrm{pcs}$; the supplementary material gives further model and implementation details.

\subsubsection{Controller} We instantiate the feedforward and feedback controllers from Sec.~\ref{subsec:physical_reservoir} as two Multi-Layer Perceptrons (MLPs):
\begin{equation*}
    \vect{T}_\mathrm{ff}(\vect{u})=\mathrm{MLP}(\vect{u}; \theta_\mathrm{contr}^\mathrm{(ff)}),
    \quad
    \vect{T}_\mathrm{fb}(\vect{Z})=\mathrm{MLP}(\vect{Z}; \theta_\mathrm{contr}^\mathrm{(fb)}).
\end{equation*}
The trainable controller parameters are $\theta_\mathrm{contr}=\{ \theta_\mathrm{contr}^\mathrm{(ff)},\theta_\mathrm{contr}^\mathrm{(fb)} \}$, i.e., the networks' weights and biases. Both MLPs use two hidden layers of 64 neurons, tanh activations, and a linear output layer.

\subsubsection{Physical--Reference Reservoir Mapping} We implement the map $\phi$ in Eq.~\eqref{eq:y2q_and_q2y} as an affine transformation parametrized through a Singular Value Decomposition (SVD) $\vect{Q}=\phi(\vect{y})=\vect{A}\vect{y}+\vect{c}$. Here, $\vect{A}=\vect{U} \bm{\Sigma} \vect{V}^\top\in\R^{3Nn_\mathrm{pcs} \times n_y}$, $\vect{c}\in\R^{3Nn_\mathrm{pcs}}$, and $s_i$ are the singular values in $\bm{\Sigma}$. If $s_i>0$ for all $i$ and $3Nn_\mathrm{pcs}=n_y$, then $\phi$ satisfies the required mapping properties from Sec.~\ref{subsec:mapping}, and its inverse is
\begin{equation*}
    \vect{y}=\phi^{-1}(\vect{Q})=\psi(\vect{Q})=\vect{A}^{-1}\left( \vect{Q}-\vect{c} \right).
\end{equation*}
This design reduces computation by dropping the second-order derivatives in Eq.~\eqref{eq:loss_fn}. The learnable mapping parameters $\theta_\mathrm{map}$ are the offset vector $\vect{c}$ and the matrix $\vect{A}$.

\subsubsection{Optimization Algorithm} We solve \eqref{eq:optimization_problem} with parallel multi-start gradient descent using Adam~\cite{adam_a_method_for_stochastic_optimization}, cosine learning-rate decay, linear warmup, and gradient clipping. Differentiable soft robot simulators~\cite{soromox_paper,soromox_package,softzoo} provide scalable parameter gradients; independent starts enable parallel execution and reduce sensitivity to local minima~\cite{rethinking_optimization_with_differentiable_simulation_etc}.

For this instantiation, $\mathcal{C}_\mathrm{phys}$ requires strictly positive geometric and physical parameters $\theta_\mathrm{phys}=\theta_\mathrm{rob}$. The mapping constraint $\mathcal{C}_\mathrm{map}$ enforces $s_i>s_\mathrm{min}=10^{-4}$ to maintain invertibility and conditioning. In all cases, lower bounds are imposed with $\bar\theta=\ln(1+\mathrm{e}^{\theta})+\theta_\mathrm{min}$.

The controller parameters are initialized with the Glorot/Xavier approach~\cite{glorot_xavier_initialization}, the mapping is initialized with $\vect{c}=\vect{0}$ and a well-conditioned SVD factorization of $\vect{A}$, and the robot parameters are sampled from predefined (physical) ranges. Specific initialization values, hardware, software, and hyperparameters are reported in the supplementary material.

For each of the 16 datasets described above, we ran four parallel trials, each initialized with a different random seed.

\begin{table}[t]
    \small
    \centering
    \setlength{\tabcolsep}{1.8pt}
    \renewcommand{\arraystretch}{1.0}
    \begin{tabular}{lccccccc}
        \toprule
        \multirow{2}{*}{\textbf{}} & \multirow{2}{*}{$\bm{n_y}$}
        & \multirow{2}{*}{$\bm{N}$} & \multirow{2}{*}{$\bm{n_\mathrm{pcs}}$}
        & \multicolumn{2}{c}{$\bm{\theta_\mathrm{rob}}$ \textbf{fixed} $(\boldsymbol{\downarrow})$}
        & \multicolumn{2}{c}{\textbf{Full pretraining} $(\boldsymbol{\downarrow})$} \\
        \cmidrule(lr){5-6} \cmidrule(lr){7-8}
        & & & & \makecell[c]{Test\\RMSE}
        & \makecell[c]{Trajectory\\NRMSE}
        & \makecell[c]{Test\\RMSE}
        & \makecell[c]{Trajectory\\NRMSE} \\
        \midrule
        \multirow{4}{*}{\rotatebox[origin=c]{90}{\textbf{sMNIST}}}
            &  6 & 1 & 2 & $2.024_{1.904}$ & $0.164_{0.157}^{**}$ & $\mathbf{0.072_{0.005}}$ & $\mathbf{0.010_{0.003}}$ \\
            &  9 & 3 & 1 & $0.282_{0.110}$ & $0.114_{0.066}$ & $\mathbf{0.079_{0.020}}$ & $\mathbf{0.010_{0.003}}$ \\
            & 12 & 2 & 2 & $4.521_{4.951}$ & $0.771_{0.785}^*$ & $\mathbf{0.132_{0.004}}$ & $\mathbf{0.025_{0.011}}$ \\
            & 15 & 5 & 1 & $5.315_{3.874}$ & $1.291_{0.338}$ & $\mathbf{0.256_{0.083}}$ & $\mathbf{0.415_{0.080}}$ \\
        \midrule
        \multirow{4}{*}{\rotatebox[origin=c]{90}{\textbf{ADIAC}}}
            &  6 & 1 & 2 & $0.370_{0.472}$ & $0.139_{0.176}$ & $\mathbf{0.072_{0.002}}$ & $\mathbf{0.048_{0.001}}$ \\
            &  9 & 3 & 1 & $0.471_{0.425}$ & $0.059_{0.058}$ & $\mathbf{0.148_{0.013}}$ & $\mathbf{0.026_{0.002}}$ \\
            & 12 & 2 & 2 & $1.842_{1.198}$ & $0.397_{0.512}$ & $\mathbf{0.711_{0.013}}$ & $\mathbf{0.053_{0.005}}$ \\
            & 15 & 5 & 1 & $1.589_{0.622}$ & $0.191_{0.121}$ & $\mathbf{1.016_{0.012}}$ & $\mathbf{0.086_{0.002}}$ \\
        \midrule
        \multirow{4}{*}{\rotatebox[origin=c]{90}{\textbf{Mac.--Gl.}}}
            &  6 & 1 & 2 & $0.181_{0.315}$ & $0.906_{1.706}$ & $\mathbf{0.010_{0.002}}$ & $\mathbf{0.048_{0.001}}$ \\
            &  9 & 3 & 1 & $0.595_{0.928}$ & $0.653_{0.196}$ & $\mathbf{0.117_{0.005}}$ & $\mathbf{0.052_{0.006}}$ \\
            & 12 & 2 & 2 & $2.348_{4.113}$ & $1.253_{0.021}^*$ & $\mathbf{0.192_{0.021}}$ & $\mathbf{0.079_{0.003}}$ \\
            & 15 & 5 & 1 & $1.137_{0.435}$ & $0.325_{0.061}^*$ & $\mathbf{0.663_{0.006}}$ & $\mathbf{0.187_{0.007}}$ \\
        \midrule
        \multirow{4}{*}{\rotatebox[origin=c]{90}{\textbf{Lorenz96}}}
            &  6 & 1 & 2 & $0.043_{0.067}$ & $0.716_{0.364}$ & $\mathbf{0.002_{0.001}}$ & $\mathbf{0.210_{0.010}}$ \\
            &  9 & 3 & 1 & $0.015_{0.020}$ & $0.550_{0.219}^{**}$ & $\mathbf{0.003_{0.001}}$ & $\mathbf{0.178_{0.065}}$ \\
            & 12 & 2 & 2 & $0.197_{0.099}$ & $0.813_{0.193}$ & $\mathbf{0.036_{0.004}}$ & $\mathbf{0.432_{0.056}}$ \\
            & 15 & 5 & 1 & $0.330_{0.209}$ & $1.061_{0.525}$ & $\mathbf{0.038_{0.002}}$ & $\mathbf{0.443_{0.034}}$ \\
        \bottomrule
    \end{tabular}
    \caption{Results of the reservoir co-optimization. For each task, four datasets from digital reference reservoirs of different dimensions $n_y$ are considered. The physical reservoir to be optimized is composed of $N$ robots represented with the PCS model with $n_\mathrm{pcs}$ segments. The results are reported as mean and standard deviation across 4 runs ($\mathrm{mean}_{\mathrm{std\; dev}}$): the RMSE on the test dataset and the NRMSE between rolled out trajectories are the metrics. For each dataset, the table also shows the outcome of the optimization where only the map and the controller were optimized ($\theta_\mathrm{rob}$ fixed). For each dataset, the best result is marked in bold. $^*$ denotes how many simulations were diverging across the 4 runs: the corresponding NRMSE were excluded from the results.}
    \label{tab:optimization_results}
\end{table}

\subsection{Optimization Results} 
\label{subsec:optimiz_results}

Tab.~\ref{tab:optimization_results} reports the optimization results (``Full pretraining'') as the mean and standard deviation, across the four random seeds, of the test Root Mean Square Error (RMSE) between the target accelerations $\vect{\ddot y}$ and the physical reservoir accelerations $\vect{\hat{\ddot y}}$. To isolate the contribution of morphology optimization, we evaluate a baseline (``$\theta_\mathrm{rob}$ fixed'') reflecting common PRC settings in which the robot morphology is treated as given: $\theta_\mathrm{rob}$ remains fixed after initialization, while only the mapping and controller parameters, $\theta_\mathrm{map}$ and $\theta_\mathrm{contr}$, are optimized. This tests whether learning the map and controller suffices to reproduce the desired dynamics or whether adapting the physical substrate is beneficial.

The results in Tab.~\ref{tab:optimization_results} show that fixing $\theta_\mathrm{rob}$ leads to larger RMSE in all scenarios. Averaging the per-$n_y$ relative reductions in test RMSE, full pretraining improves over the fixed-morphology baseline by \SI{90.2}{\percent} for sMNIST, \SI{61.6}{\percent} for ADIAC, \SI{77.1}{\percent} for Mackey--Glass, and \SI{86.4}{\percent} for Lorenz96. This indicates that morphology optimization is important for accurately matching the target digital dynamics. The fixed-morphology baseline also exhibits substantially larger standard deviations, showing that its outcome is more sensitive to initialization. In contrast, full pretraining yields low standard deviations across the tested settings, indicating consistent convergence across random seeds.

The test RMSE generally increases with the reservoir dimension $n_y$. This is expected, since larger reservoirs induce more complex target dynamics that are harder for the physical reservoir to approximate precisely. Moreover, since the dataset dimension $m$ is kept fixed while the phase-space dimension of $(\vect{y},\vect{\dot y})$ grows with $n_y$, the training data become progressively sparser in this space, which can further make the dynamics-matching problem harder.

\begin{figure}[!t]
    \centering
    \includegraphics[width=0.35\columnwidth]{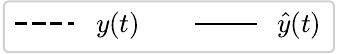}
    \\[0.3em]
    \begin{subfigure}{0.47\columnwidth}
        \centering
        \includegraphics[width=\textwidth]{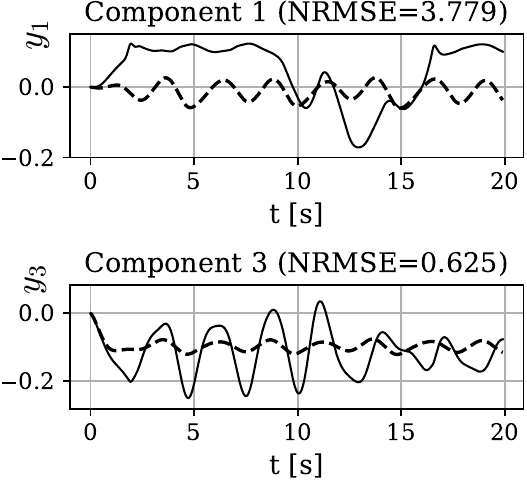}
    \caption{Co-optimization of map and controllers ($\theta_\mathrm{map}$ and $\theta_\mathrm{contr}$). Soft robots ($\theta_\mathrm{rob}$) are held fixed.}
    \label{fig:optimization_simulation_example_no_robots}
    \end{subfigure}
    \hfill
    \begin{subfigure}{0.47\columnwidth}
        \centering
        \includegraphics[width=\textwidth]{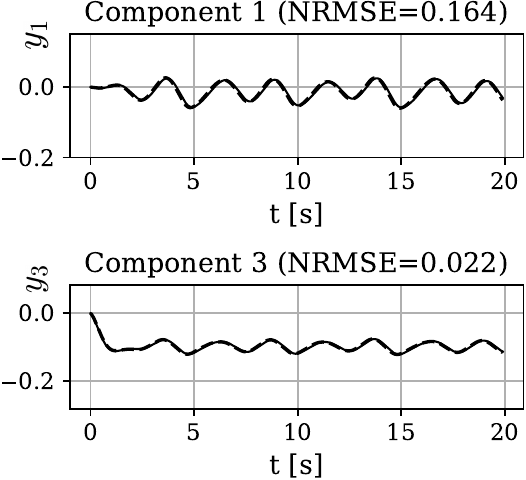}
    \caption{``Full'' co-optimization of robots, map and controllers ($\theta_\mathrm{rob}$, $\theta_\mathrm{map}$ and $\theta_\mathrm{contr}$).}
    \label{fig:optimization_simulation_example_full}
    \end{subfigure}
    \caption{Time evolution of the mapped physical coordinates $\vect{\hat y}$ (solid line) and reference coordinates $\vect{y}$ (dashed line) after optimization with the Mackey--Glass $n_y=6$ dataset. Only two components ($y_1$ and $y_3$) are shown.}
    \label{fig:optimization_simulation_example}
\end{figure}

Beyond the acceleration-based test RMSE, we also evaluate the pretrained reservoirs by simulating them under a given input $\vect{u}(t)$ and comparing their response $\vect{\hat y}(t)$ with the corresponding digital reservoir trajectory $\vect{y}(t)$. Fig.~\ref{fig:optimization_simulation_example} shows a representative Mackey--Glass example for $n_y=6$, comparing full pretraining with the fixed-morphology baseline, while Tab.~\ref{tab:optimization_results} reports the Normalized Root Mean Square Error (NRMSE) between $\vect{\hat y}(t)$ and $\vect{y}(t)$, i.e., the RMSE divided by the root-mean-square value of $\vect{y}(t)$. Consistently with the RMSE results, optimizing only the map and MLP controllers is not sufficient to reproduce the target dynamics reliably. In some fixed-morphology runs (7 out of 64), the final simulated system also becomes unstable for the given input: these diverging runs are marked with an asterisk in Tab.~\ref{tab:optimization_results} and do not contribute to the mean and standard deviation computation.

Overall, the proposed pretraining procedure produces physical reservoirs that closely approximate the target digital dynamics. This behavior is consistent across all tested datasets and random starts.

\section{Physical Reservoir Computing Experiments}
\label{sec:benchmarking}

This section evaluates whether the dynamics-matching pretraining from Sec.~\ref{sec:optimization_experiments} yields physical reservoirs that improve downstream PRC performance. The pretrained reservoirs are benchmarked through simulations on standard temporal Machine Learning (ML) tasks and compared with controlled PRC baselines.

\subsection{Benchmarking Setup} 
\label{subsec:benchmarking_setup}

The benchmark tasks are sMNIST and ADIAC classification, and Mackey--Glass and Lorenz96 time-series forecasting. For each task, we test the four pretrained reservoir dimensions $n_y\in\{6,9,12,15\}$ in the full PRC architecture: the soft robot, map, and controller obtained in Sec.~\ref{subsec:optimiz_results} are fixed as the physical reservoir and mapping in Fig.~\ref{fig:methodology_high_level}, and only the task-specific readout is trained. Performance is measured by classification accuracy [\%] for sMNIST and ADIAC, and NRMSE [-] for Mackey--Glass and Lorenz96, which is the test RMSE divided by the root-mean-square value of the test sequence. These metrics follow the RON evaluation in~\cite{random_oscillators_network_for_time_series_processing}. Dataset and readout-training details are provided in the supplementary material.

\begin{figure*}[t]
    \centering
    \includegraphics[width=0.55\textwidth]{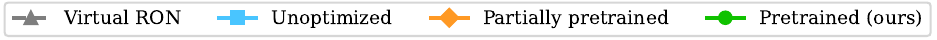}
    \\[0.3em]
    \begin{subfigure}{0.24\textwidth}
        \centering
        \includegraphics[width=\textwidth]{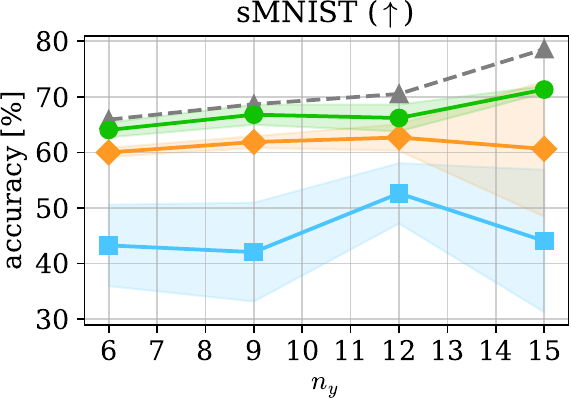}
        \caption{sMNIST task. Accuracy as \% of correct predictions.}
        \label{fig:smnist_prc_scalability}
    \end{subfigure}
    \hfill
    \begin{subfigure}{0.24\textwidth}
        \centering
        \includegraphics[width=\textwidth]{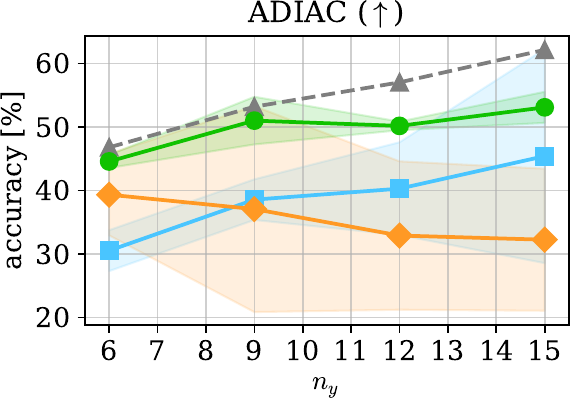}
        \caption{ADIAC task. Accuracy as \% of correct predictions.}
        \label{fig:adiac_prc_scalability}
    \end{subfigure}
    \hfill
    \begin{subfigure}{0.24\textwidth}
        \centering
        \includegraphics[width=\textwidth]{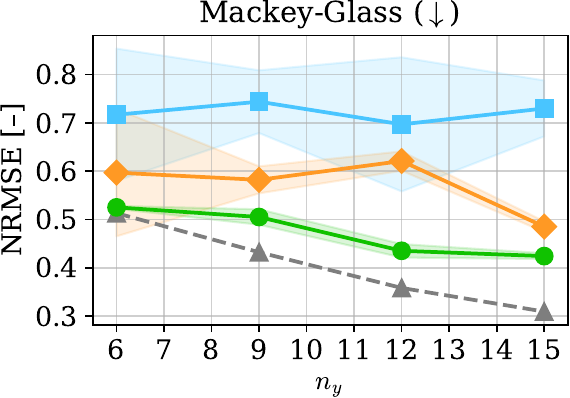}
        \caption{Mackey--Glass task. Prediction error as NRMSE.}
        \label{fig:mackey_prc_scalability}
    \end{subfigure}
    \hfill
    \begin{subfigure}{0.24\textwidth}
        \centering
        \includegraphics[width=\textwidth]{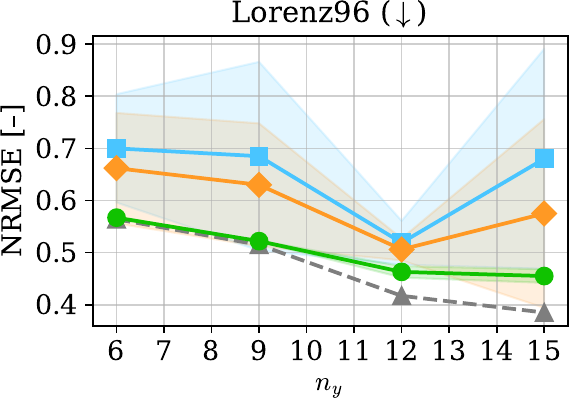}
        \caption{Lorenz96 task. Prediction error as NRMSE.}
        \label{fig:lorenz_prc_scalability}
    \end{subfigure}
    \caption{Scalability of the results of the PRC benchmark experiments. Each plot shows the resulting performance vs the reservoir dimension $n_y$. Performance is evaluated as the mean (line) and standard deviation (shaded area) across 4 different runs. The reference RON performance is plotted as a dashed gray line.}
    \label{fig:prc_scalability}
\end{figure*}

We compare the proposed reservoir with one digital reference and two physical PRC baselines:
\begin{enumerate}[label=\alph*)]
    \item \textit{Digital RON reference}: the nominal reservoir used as the pretraining target. This reference quantifies how much of its task performance is retained after physical realization.
    \item \textit{Unoptimized PRC baseline}: a conventional PRC system in which the physical substrate is used without pretraining, map, or feedback controller. This baseline measures the benefit of pretraining relative to a substrate-as-given PRC implementation.
    \item \textit{Partially pretrained baseline}: a reservoir in which the soft robot parameters are fixed and only the map and controllers are optimized. This baseline isolates the contribution of optimizing the physical morphology.
    \item \textit{Pretrained (ours)}: the physical reservoir and mapping obtained by the proposed full pretraining procedure.
\end{enumerate}
In the \textit{unoptimized PRC baseline}, the input $\vect{u}(t)$ actuates the soft robots through a randomly initialized nonlinear encoding
\begin{equation}
   \vect{T}=\tanh(\vect{Vu}+\vect{d}),
   \label{eq:random_encoding}
\end{equation}
with $\vect{V}\in\R^{3Nn_\mathrm{pcs}\times n_u}$ and $\vect{d}\in\R^{3Nn_\mathrm{pcs}}$.

The \textit{partially pretrained baseline} uses the reservoirs from Sec.~\ref{sec:optimization_experiments} in which $\theta_\mathrm{rob}$ was held fixed. Sec.~\ref{subsec:optimiz_results} showed that this setting does not consistently approximate the nominal target dynamics. Evaluating it here determines whether optimizing only the map and controller can nevertheless provide sufficient expressiveness, memory, and task-relevant dynamics for competitive PRC performance.

For statistical comparison, each experiment is repeated four times. The \textit{pretrained (ours)} and \textit{partially pretrained} cases use the four reservoirs obtained from the four starts of the parallel multi-start gradient descent procedure in Sec.~\ref{sec:optimization_experiments}. The \textit{unoptimized PRC baseline} uses four random initializations of the robot parameters and of $\vect{V},\vect{d}$ in Eq.~\ref{eq:random_encoding}.

\subsection{Benchmarking Results}
\label{subsec:benchmarking_results}

\begin{table}[!t]
    \small
    \centering
    \setlength{\tabcolsep}{2pt}
    \renewcommand{\arraystretch}{1.0}
    \begin{tabular}{llcccc}
        \toprule
        \textbf{Task} & $\bm{n_y}$
        & \cellcolor{gray!15}\makecell[c]{\textbf{Digital}\\ \textbf{Reserv.}}
        & \makecell[c]{\textbf{Unopti-}\\ \textbf{mized}}
        & \makecell[c]{\textbf{Partially}\\ \textbf{pretrained}}
        & \makecell[c]{\textbf{Pretrained}\\ \textbf{(ours)}} \\
        \midrule
        \multirow{4}{*}{\makecell[l]{sMNIST\\(\% $\uparrow$)}}
            &  6 & \cellcolor{gray!15}$65.88$ & $43.28_{7.32}$  & $59.99_{0.83}$  & $\mathbf{64.03_{1.32}}$ \\
            &  9 & \cellcolor{gray!15}$68.66$ & $42.07_{8.89}$  & $61.86_{1.06}$  & $\mathbf{66.78_{1.82}}$ \\
            & 12 & \cellcolor{gray!15}$70.51$ & $52.63_{5.43}$  & $62.68_{2.39}$  & $\mathbf{66.18_{2.41}}$ \\
            & 15 & \cellcolor{gray!15}$78.57$ & $44.08_{12.79}$ & $60.61_{12.16}$ & $\mathbf{71.30_{0.63}}$ \\
        \midrule
        \multirow{4}{*}{\makecell[l]{ADIAC\\(\% $\uparrow$)}}
            &  6 & \cellcolor{gray!15}$46.80$ & $30.54_{3.22}$  & $39.35_{6.41}$  & $\mathbf{44.58_{1.03}}$ \\
            &  9 & \cellcolor{gray!15}$53.20$ & $38.57_{3.20}$  & $37.08_{16.21}$  & $\mathbf{51.01_{3.75}}$ \\
            & 12 & \cellcolor{gray!15}$57.03$ & $40.30_{7.30}$  & $32.92_{11.69}$  & $\mathbf{50.18_{0.68}}$ \\
            & 15 & \cellcolor{gray!15}$62.15$ & $45.42_{16.85}$ & $32.26_{11.18}$ & $\mathbf{53.10_{2.45}}$ \\
        \midrule
        \multirow{4}{*}{\makecell[l]{Mackey--Glass\\(NRMSE $\downarrow$)}}
            &  6 & \cellcolor{gray!15}$0.513$ & $0.717_{0.137}$ & $0.597_{0.132}$   & $\mathbf{0.525_{0.003}}$ \\
            &  9 & \cellcolor{gray!15}$0.432$ & $0.744_{0.065}$ & $0.582_{0.028}$   & $\mathbf{0.505_{0.016}}$ \\
            & 12 & \cellcolor{gray!15}$0.358$ & $0.697_{0.139}$ & $0.621_{0.020}^*$ & $\mathbf{0.435_{0.014}}$ \\
            & 15 & \cellcolor{gray!15}$0.309$ & $0.730_{0.058}$ & $0.485_{0.011}^*$ & $\mathbf{0.425_{0.006}}$ \\
        \midrule
        \multirow{4}{*}{\makecell[l]{Lorenz96\\(NRMSE $\downarrow$)}}
            &  6 & \cellcolor{gray!15}$0.564$ & $0.699_{0.104}$ & $0.662_{0.106}$ & $\mathbf{0.567_{0.002}}$ \\
            &  9 & \cellcolor{gray!15}$0.515$ & $0.685_{0.180}$ & $0.630_{0.118}$ & $\mathbf{0.522_{0.001}}$ \\
            & 12 & \cellcolor{gray!15}$0.417$ & $0.519_{0.042}$ & $0.506_{0.021}$ & $\mathbf{0.463_{0.011}}$ \\
            & 15 & \cellcolor{gray!15}$0.385$ & $0.680_{0.211}$ & $0.575_{0.181}$ & $\mathbf{0.455_{0.013}}$ \\
        \bottomrule
    \end{tabular}
    \caption{Results of the PRC benchmark experiments on simulated physical reservoir rollouts. For each task and reservoir dimension ($n_y$), the pretrained physical reservoir is compared with unoptimized PRC structures and the digital RON reference. Results are mean and standard deviation ($\mathrm{mean}_{\mathrm{std\; dev}}$) over 4 seeds of classification accuracy for sMNIST and ADIAC, and NRMSE for Mackey--Glass and Lorenz96. The best physical architecture is highlighted in bold. $^*$ denotes one diverging trial, which is excluded from the results.}
    \label{tab:benchmark_result}
\end{table}

Tab.~\ref{tab:benchmark_result} reports the benchmark results, and Fig.~\ref{fig:prc_scalability} shows the same data as a function of $n_y$. Across all task-dimension pairs, the fully pretrained reservoir is the best-performing physical PRC configuration. To aggregate results across the mixed metrics, we measure relative performance by comparing accuracy values directly for sMNIST and ADIAC, where higher values are better, and by comparing the inverse of the error for Mackey--Glass and Lorenz96, where lower NRMSE values are better. Baseline improvements are computed in the same direction: as the relative accuracy gain for sMNIST and ADIAC, and as the relative NRMSE reduction for the two forecasting tasks.

Averaged uniformly over the 16 task-dimension cases in Tab.~\ref{tab:benchmark_result}, the optimized physical reservoir reaches \SI{90.9}{\percent} of the digital RON reference performance. The same normalization shows that full pretraining improves performance over the \textit{unoptimized PRC baseline} by \SI{33.7}{\percent} on average, using accuracy gain for sMNIST and ADIAC, and NRMSE reduction for Mackey--Glass and Lorenz96. Relative to the \textit{partially pretrained baseline}, the average improvement is \SI{19.2}{\percent}. These gains indicate that learning the map and controller alone improves, in general, over an unoptimized substrate, but optimizing the physical reservoir is required to approach the target RON performance. The unoptimized baseline also has substantially larger standard deviations, showing that its performance is more sensitive to random initialization.

The gap between the fully pretrained reservoir and the digital RON reference generally increases with $n_y$. This trend is consistent with Sec.~\ref{subsec:optimiz_results}, where larger reservoirs yielded higher pretraining RMSE and therefore less accurate approximations of the nominal RON dynamics. At the same time, increasing the optimized physical reservoir from $n_y=6$ to $n_y=15$ improves task performance by \SI{17.3}{\percent} on average: accuracy increases by \SI{11.4}{\percent} for sMNIST and by \SI{19.1}{\percent} for ADIAC, while NRMSE decreases by \SI{19.0}{\percent} for Mackey--Glass and \SI{19.8}{\percent} for Lorenz96.

A digital inference-cost comparison against the RON is provided in the supplementary material, showing that although the physical reservoir still requires digital computation for its controllers, with the affine map absorbed into the readout, this cost scales linearly rather than quadratically with reservoir dimension.

The task-specific RON tuning used here (e.g., spectral radius and time step) follows established practice~\cite{echo_state_network_tutorial,exploration_of_effects_of_different_network_topologies_on_esn,optimization_of_leaky_esn,esn_with_band_pass_neurons,random_oscillators_network_for_time_series_processing} and yields task-specialized physical reservoirs; supplementary experiments nevertheless show effective cross-task reuse without morphology reoptimization.

\section{Conclusions}
\label{sec:conclusions}

We presented a pretraining method for soft robotic physical reservoirs that co-optimizes morphology, a physical--reference state map, and a controller to approximate digital reference dynamics. The method uses a dynamics-level equation-error objective, avoiding task-level optimization during pretraining.
Using a Random Oscillators Network (RON)~\cite{random_oscillators_network_for_time_series_processing} as the reference model, the proposed approach reduced acceleration-matching errors and the optimized physical reservoirs approximate closely the reference dynamics across the tested datasets. Comparisons with fixed-morphology baselines indicate that morphology optimization is a key contributor to this approximation.
When evaluated in simulated full PRC pipelines on sMNIST, ADIAC, Mackey--Glass, and Lorenz96 tasks, the pretrained reservoirs outperformed unoptimized and partially pretrained physical baselines and retained \SI{90.9}{\percent} of the digital RON reference performance on average.
%
These preliminary results support dynamics-level co-optimization for low-dimensional soft robotic PRC. For computational tractability, the reference RONs and corresponding physical reservoirs were restricted to $n_y\in\{6,9,12,15\}$, which limits absolute task performance relative to state-of-the-art digital RC approaches; scaling the method to higher-dimensional physical reservoirs is therefore a primary direction for future work. Further work should validate the method on hardware and explicitly incorporate unmodeled dynamics, substrate variability, and sensor noise during pretraining and co-optimization to improve robustness to discrepancies between simulated and physical behavior. Another research direction is to characterize the compatibility between digital and physical reservoir vector fields, identifying which target dynamics are reachable by a given class of physical substrates. It will also be important to investigate the trade-off between the complexity of the physical--reference map and controllers and the computational contribution of the body itself, since a highly expressive map or controller could compensate for morphological limitations at the cost of a higher digital overhead.


\section*{Acknowledgments}
The work by M. Stölzle, M. Ramírez Montero, and Cosimo Della Santina was supported under the European Union's Horizon Europe Program from Project EMERGE - Grant Agreement No. 101070918. 
The work by M. Stölzle and D. Rus was supported in part by the Singapore MIT Alliance on Research and Technology (SMART) under the Mens, Manus, et Machina (M3S) program, by the BARI EMERGE program under grant N00014-26-1-2304, and by the MIT-GIST collaboration.
The work of C. Della Santina was further supported by the GRAIL project.

\bibliography{aaai2027}

\def\isSupplementIncluded{}
\makeatletter
\@ifundefined{isSupplementIncluded}{%
  \newif\ifSupplementStandalone
  \SupplementStandalonetrue
}{%
  \newif\ifSupplementStandalone
  \SupplementStandalonefalse
}
\makeatother

\ifSupplementStandalone
  \newcommand{\FinishSupplementDocument}{%
    \newpage
    \bibliography{aaai2027}
    \end{document}%
  }
  \documentclass[11pt,letterpaper]{article}
  \usepackage[margin=1in]{geometry}
  \usepackage[T1]{fontenc}
  \usepackage{newtxtext}

\usepackage[hyphens]{url}
\usepackage{graphicx}
\urlstyle{rm}
\def\UrlFont{\rm}

\usepackage{natbib}
\usepackage{caption}
\usepackage{subcaption}
\frenchspacing

\usepackage{booktabs}
\usepackage{amsmath}
\usepackage{amssymb}
\usepackage{comment}
\usepackage{bm}
\usepackage{mathrsfs}
\usepackage[cal=cm]{mathalpha}
\usepackage{makecell}
\usepackage{todonotes}
\usepackage{enumitem}
\usepackage{colortbl}
\usepackage{xcolor}
\usepackage{siunitx}
\usepackage{multirow}

\definecolor{lightgray}{gray}{0.85}
\newcommand{\vect}[1]{\mathbf{#1}}
\newcommand{\R}{\mathbb{R}}
\newcommand{\K}{\mathord{\scalebox{0.8}{$\mathcal{K}$}}}

\setcounter{secnumdepth}{2}

  \captionsetup{font=small,labelfont=bf}
  \let\cite\citep
  \let\shortcite\citeyearpar
  \setcitestyle{aysep={}}
  \setlength\bibhang{0pt}
  \bibliographystyle{aaai2027}
  \let\endthebibliography=\endlist

  \title{\vspace{-2em}\huge Supplementary Material}
  \date{}
  \author{}

  \begin{document}
  \maketitle
  \vspace{-4em}
  \thispagestyle{empty}
  \pagestyle{empty}
\else
  \newcommand{\FinishSupplementDocument}{}
  \clearpage
  \onecolumn
  \begin{center}
    {\LARGE\bfseries Supplementary Material}
  \end{center}
  \vspace{1em}
\fi

\appendix

\section{Physical--Reference Reservoir Mapping}
\label{app:mapping}
This appendix provides a more detailed discussion about the transformation used to map coordinates between the digital reference space and the physical reservoir space. Let $\vect{y}(t)\in\R^{n_y}$ denote the digital reference coordinates and $\vect{Q}(t)\in\R^{n_Q}$ the physical reservoir coordinates.
The map $\phi:\R^{n_y}\rightarrow \R^{n_Q}$ from the reference to the physical reservoir space is defined as
\begin{equation*}\label{eq:y2q}
    \vect{Q}=\phi(\vect{y}),
\end{equation*}
while the inverse transformation is
\begin{equation*}
    \vect{y}=\phi^{-1}(\vect{Q})=\psi(\vect{Q}).
\end{equation*}
To support the proposed optimization strategy, the map must be invertible for consistent transformations between $\vect{Q}$ and $\vect{y}$ and sufficiently smooth to transform velocities and accelerations. In practice, this requires a second-order smooth diffeomorphism.

We use Fr\'echet-derivative notation: $\mathrm{D}\phi(\vect{y})$ and $\mathrm{D}\psi(\vect{Q})$ denote first derivatives, interpreted as linear maps, while $\mathrm{D}^2\phi(\vect{y})$ and $\mathrm{D}^2\psi(\vect{Q})$ denote second derivatives, interpreted as symmetric bilinear maps. Then velocities and accelerations transform as
\begin{equation*}
    \dot{\vect{Q}}=\mathrm{D}\phi(\vect{y})[\dot{\vect{y}}],
    \qquad
    \dot{\vect{y}}=\mathrm{D}\psi(\vect{Q})[\dot{\vect{Q}}],
    \label{eq:yd2qd_and_qd2yd}
\end{equation*}
and
\begin{equation*}
    \ddot{\vect{Q}}=\mathrm{D}\phi(\vect{y})[\ddot{\vect{y}}]+\mathrm{D}^2\phi(\vect{y})[\dot{\vect{y}},\dot{\vect{y}}],
    \qquad
    \displaystyle
    \ddot{\vect{y}}=\mathrm{D}\psi(\vect{Q})[\ddot{\vect{Q}}]+\mathrm{D}^2\psi(\vect{Q})[\dot{\vect{Q}},\dot{\vect{Q}}].
\end{equation*}
For example, $\mathrm{D}\phi(\vect{y}):\R^{n_y}\rightarrow\R^{n_Q}$ maps a reference velocity to a physical reservoir velocity, while $\mathrm{D}^2\phi(\vect{y}):\R^{n_y}\times\R^{n_y}\rightarrow\R^{n_Q}$ maps two reference velocities to the corresponding second-order chain-rule correction. In coordinates, for $\vect{v},\vect{w}\in\R^{n_y}$ and $i=1,\ldots,n_Q$,
\begin{align*}
    \bigl(\mathrm{D}\phi(\vect{y})[\vect{v}]\bigr)_i
    &=
    \sum_{j=1}^{n_y}
    \frac{\partial\phi_i}{\partial y_j}(\vect{y})\,v_j,\\
    \bigl(\mathrm{D}^2\phi(\vect{y})[\vect{v},\vect{w}]\bigr)_i
    &=
    \sum_{j=1}^{n_y}\sum_{k=1}^{n_y}
    \frac{\partial^2\phi_i}{\partial y_j\partial y_k}(\vect{y})\,v_jw_k.
\end{align*}
Analogous expressions hold for $\psi$.

The mapping complexity determines how reference coordinates are projected into the physical reservoir space. More expressive transformations (e.g., nonlinear ones) can improve dynamics matching during optimization but require greater computational effort in the digital domain. An overly complex map may therefore shift computation away from the physical substrate, reducing its contribution.

\section{Benchmark Tasks for Physical Reservoir Computing}
\label{app:tasks}
The proposed Physical Reservoir Computing (PRC) strategy is evaluated on the sMNIST, ADIAC, Mackey--Glass and Lorenz96 benchmark tasks. These tasks were selected because they constitute standard benchmarks for Machine Learning (ML) models operating on temporal sequences. This appendix summarizes the training procedures used for the Reservoir Computing (RC) and PRC systems on each task and provides additional details on the corresponding datasets. In all cases, simulations of the physical reservoir were performed with the Diffrax library~\cite{diffrax}, using an integration step $\delta t= \SI{0.1}{ms}$ and the forward Euler method.
\\
\\
\noindent \textbf{sMNIST (sequential--MNIST)} is a classification task in which the model is required to recognize a handwritten digit from an image presented as a temporal sequence of pixels~\cite{sMNIST_paper}. It is based on the MNIST dataset~\cite{MNIST_paper}, which contains 70000 $28\times28$ grayscale images (60000 in the training set and 10000 in the test set) of handwritten digits from 0 to 9. In sMNIST, each image is fed to the reservoir as a sequence of 784 pixels, so the input $u(k)\in\R$ is a real number in $[0, \,1]$ representing the intensity of the $k$-th pixel. The model is then required to predict the corresponding digit.

\begin{figure}[!t]
    \centering
    \includegraphics[width=\textwidth]{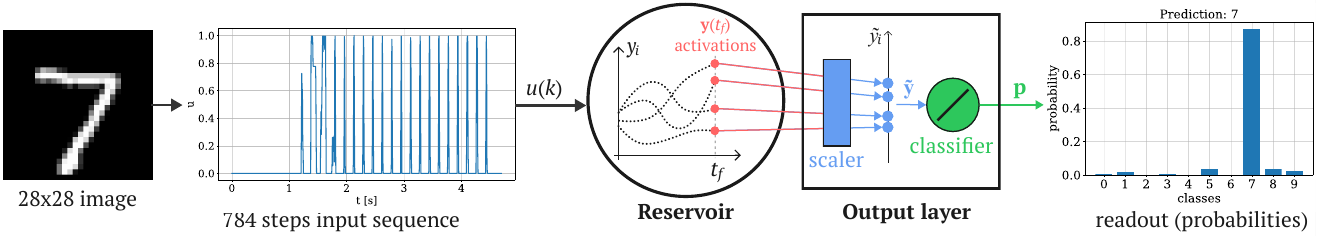}
    \caption{Schematic of how RC is used to perform the sMNIST classification task.}
    \label{fig:sMNIST_scheme}
\end{figure}

For output layer training in RC (or PRC), the input sequence is passed through the reservoir from $t_0=0$ to $t_f$, and the final reservoir coordinates $\vect{y}(t_f)\in\R^{n_y}$ are used as activations. This procedure is repeated for all $m$ images in the training set, producing a batched collection of activations $\vect{Y}(t_f)\in\R^{m\times n_y}$. First, a \textit{scaler} function is defined from the mean $\bm{\mu}\in\R^{n_y}$ and standard deviation $\bm{\sigma}\in\R^{n_y}$ of these activations over the $m$ samples; it is then applied to scale the activations as
\begin{equation*}
    \vect{\tilde{y}}_j=\mathrm{scaler}(\vect{y}_j(t_f))=\operatorname{diag}(\bm{\sigma})^{-1} \left( \vect{y}_j(t_f)-\bm{\mu} \right) \qquad\text{$\forall j=1,...,m$,}
\end{equation*}
yielding the scaled activation matrix $\vect{\tilde{Y}}\in\R^{m\times n_y}$.
A \textit{classifier} is subsequently trained on these scaled activations to estimate the probabilities of the 10 classes, corresponding to digits from 0 to 9. In particular, the classifier is implemented as a multiclass logistic regression model
\begin{equation*}
    \vect{p}=\mathrm{softmax}(\vect{W}_o\,\vect{\tilde{y}}+\vect{b}_o),
\end{equation*}
whose weights $\vect{W}_o\in\R^{10\times n_y}$ and biases $\vect{b}_o\in\R^{10}$ are computed by minimizing cross-entropy loss with L2 regularization.

During inference, an input sequence (image) is passed to the reservoir, and the activations $\vect{y}(t_f)$ are scaled using the scaler defined during training. The scaled activations $\vect{\tilde{y}}$ are then provided to the trained classifier, which outputs the predicted probabilities $\vect{p}\in\R^{10}$ for all classes. The predicted class is obtained by selecting the class with maximum probability. A schematic of this inference process is shown in Fig.~\ref{fig:sMNIST_scheme}.

The performance of the model on this task is evaluated through classification accuracy, computed as the number of correct predictions divided by the total number of images. For computational tractability, the experiments used a training set of 30000 images and a test set of 5000 images extracted from the original MNIST dataset.
\\
\\
\noindent \textbf{ADIAC (Automatic Diatom Identification And Classification)} involves automatic identification of diatoms (unicellular algae) on the basis of images that are converted into temporal sequences~\cite{adiac_task}. It is a classification task with 37 classes, and the dataset is composed of 390 train samples and 391 test samples, each consisting of a 176-steps univariate time series~\cite{adiac_dataset}. The (physical) reservoir computing system operates on this task exactly as in sMNIST.
\\
\\
\noindent \textbf{Mackey--Glass} is another standard benchmark task, in which the model predicts a time sequence generated by the Mackey--Glass equation~\cite{mackey-glass_equation}:
\begin{equation}
    \dot{u}(t)=\frac{\beta\theta^n u(t-\tau)}{\theta^n+u(t-\tau)^n}-\gamma\,u(t).
    \label{eq:MG}
\end{equation}
This system displays complex behavior for selected parameter values and is commonly used to evaluate the memory capacity of recurrent networks. The dataset is a sequence $\vect{u}=[u(0), \; ... \; , \; u(N-1)]^\top\in\R^{N}$ of $N$ values from a discrete solution of Eq.~\eqref{eq:MG}. At time $k$, the model is required to predict the future state of the system after a lag $N_l$, namely $u(k+N_l)$.

\begin{figure}[t]
    \centering
    \includegraphics[width=\textwidth]{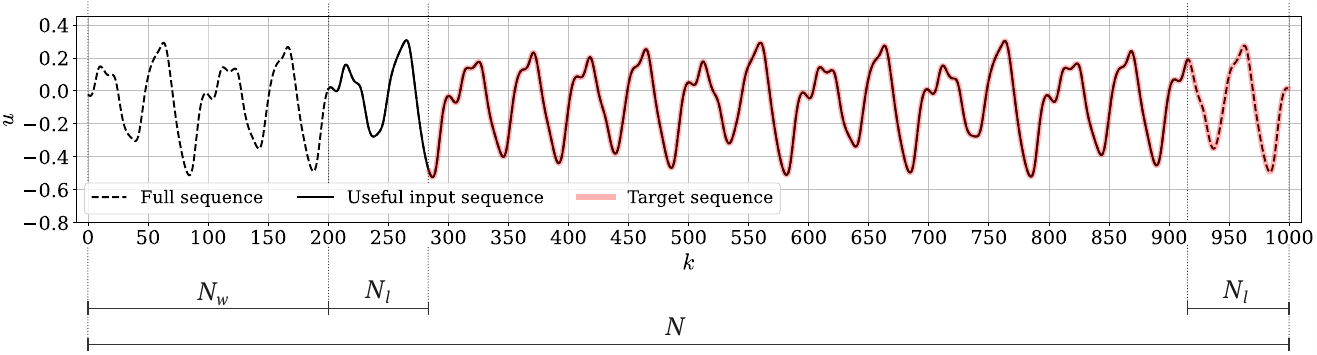}
    \caption{Visualization of Mackey--Glass dataset-sequence partition. In this specific example $N=1000,\,N_w=200,\,N_l=84$.}
    \label{fig:MG_ex}
\end{figure}

During training, an input sequence (i.e., the portion of the dataset from $k=0$ to $k=N-N_l-1$) is fed to the reservoir, and its response is recorded. The first $N_w$ steps are discarded as \textit{washout} of the initial transient, and the reservoir coordinates $\vect{y}(k)\in\R^{n_y}$ from $k=N_w$ to $k=N-N_l-1$ are used as activations. These activations are rescaled as in the sMNIST task and then used to train a linear layer
\begin{equation*}
    \hat{u}(k+N_l)=\vect{W}_o\;\vect{\tilde{y}}(k)+b_o \qquad \text{(with $\vect{W}_o\in\R^{1\times n_y}$, $b_o\in\R$)}
    \label{eq:linear_MG}
\end{equation*}
using Ridge regression. The corresponding labels, which are compared with the predictions $\hat{u}$, are the portion of the full dataset sequence from $k=N_w+N_l$ to $k=N-1$. A visual schematic of this dataset partition is shown in Fig.~\ref{fig:MG_ex}.

During inference, the prediction is obtained from the trained output layer, with the activations $\vect{\tilde y}(k)$ computed rescaling $\vect{y}(k)$ with the scaler defined during training.

The performance of the model on this task is evaluated using the Normalized Root Mean Square Error (NRMSE) between the predicted sequence $\vect{\hat{u}}\in\R^{N-N_w-N_l}$ and the target sequence $\vect{u}_\mathrm{targ}\in\R^{N-N_w-N_l}$:
\begin{equation*}
    \mathrm{NRMSE}(\vect{u}_\mathrm{targ},\vect{\hat{u}}) = \frac{\mathrm{RMSE}(\vect{u}_\mathrm{targ},\vect{\hat{u}})}{\mathrm{RMS}(\vect{u}_\mathrm{targ})} = \frac{\sqrt{\frac{1}{N-N_w-N_l}\,\lVert \vect{u}_\mathrm{targ} - \vect{\hat{u}} \rVert_2^2}}{\sqrt{\frac{1}{N-N_w-N_l}\,\lVert \vect{u}_\mathrm{targ} \rVert_2^2}}.
\end{equation*}
In our experiments we adopted a prediction lag $N_l$ of 84 steps and $N_w=200$ washout steps for the initial transient, with one training and one test temporal sequence of $N=2000$ steps each.
\\
\\
\noindent \textbf{Lorenz96} is a time-series forecasting task in which, at time $k$, the model is required to predict the future state $\vect{u}(k+N_l)\in\R^{n_u}$ of a temporal sequence generated by the multidimensional Lorenz equation~\cite{lorenz96}:
\begin{equation*}
    \begin{cases}
        \dot u_i=(u_{i+1}-u_{i-2}) u_{i-1}-u_i+F\\
        u_{-1}=u_{n_u-1},\; u_0=u_{n_u},\; u_{n_u+1}=u_1
    \end{cases}
     \qquad (\text{with $i=1,\, ...\, , n_u$}).
\end{equation*}
For this task, the model is trained analogously to the Mackey--Glass case, except that the input has dimension $n_u > 1$. In our experiments, $n_u=5$ and $F=8$. The evaluation metric is the NRMSE between the target and predicted sequences. A prediction lag $N_l$ of 25 steps, $N_w=200$ washout steps, and 128 batched sequences of $N=2225$ steps each were used for both training and testing.

\section{Digital Reference Dynamics and Pretraining Dataset Generation}
\label{app:dataset_collection}
The proposed co-optimization methodology requires appropriate digital reservoir dynamics to serve as the reference. We use the Random Oscillators Network (RON) introduced in~\cite{random_oscillators_network_for_time_series_processing} because it performs well on several tasks and is inspired by nonlinearly coupled physical oscillators. Its dynamics are therefore compatible with the soft robot. This appendix introduces the RON and describes the datasets of points $(\vect{y},\dot{\vect{y}},\vect{u})$ and labels $\ddot{\vect{y}}$ used for pretraining.

\paragraph{Random Oscillators Network model} The RON is a digital reservoir composed of $n_y$ mass-spring-damper elements coupled by a nonlinear forcing term. Its continuous-time equations of motion are:
\begin{equation}
    \ddot{\vect{y}}=-\bm{\gamma}\odot\vect{y}-\bm{\varepsilon}\odot\dot{\vect{y}}+\tanh(\vect{W}\vect{y}+\vect{V}\vect{u}+\vect{b}).
    \label{eq:RON_equation}
\end{equation}
Here, $\vect{y}(t)\in\R^{n_y}$ collects the oscillator positions (reservoir coordinates), $\vect{u}(t)\in\R^{n_u}$ is the input vector, $\bm{\gamma}\in\R^{n_y}$ and $\bm{\varepsilon}\in\R^{n_y}$ collect the stiffness and damping factors, respectively, $\vect{W}\in\R^{n_y\times n_y}$ is the hidden-to-hidden weight matrix, $\vect{V}\in\R^{n_y\times n_u}$ is the input-to-hidden weight matrix, $\vect{b}\in\R^{n_y}$ is the bias vector, and $\odot$ denotes the element-wise product.
In state-space form, calling $\vect{z}=[\vect{y}, \; \dot{\vect{y}}]^\top$:
\begin{equation} \label{eq:RON_ss}
    \begin{cases}
        \dot{\vect{z}}_1=\vect{z}_2 \\
        \dot{\vect{z}}_2=-\bm{\gamma} \odot \vect{z}_1 - \bm{\varepsilon} \odot \vect{z}_2+\tanh{(\vect{W}\vect{z}_1+\vect{V}\vect{u}+\vect{b})}.
    \end{cases}
\end{equation}
Eq.~\eqref{eq:RON_ss} is then discretized with temporal step $\Delta t$ to obtain the discrete-time equations of the digital reservoir, using implicit Euler method for $\vect{z}_1$ equation and explicit Euler method for $\vect{z}_2$ equation:
\begin{equation} \label{eq:RON_discrete}
    \begin{cases}
        \vect{z}_1(k+1)=\vect{z}_1(k)+\Delta t \; \vect{z}_2(k+1) \\
        \vect{z}_2(k+1)=\vect{z}_2(k) + \Delta t \; \left[ -\bm{\gamma} \odot \vect{z}_1(k) - \bm{\varepsilon} \odot \vect{z}_2(k)+\tanh{(\vect{W}\vect{z}_1(k)+\vect{V}\vect{u}(k)+\vect{b})} \right].
    \end{cases}
\end{equation}
The spectral radius of $\vect{W}$ is denoted by $\rho_\mathrm{w}$, and $\nu$ denotes the input scaling factor used to rescale $\vect{V}$ and $\vect{b}$ at initialization.

\paragraph{Reference RON configuration selection} The properties of a RON reservoir, including memory retention, expressiveness, and timescale, depend on its \textit{configuration}, i.e., the values of its parameters $n_y, \bm{\gamma}, \bm{\varepsilon}, \vect{W}, \vect{V}, \vect{b}$, $\Delta t$ and $\nu$. In~\cite{random_oscillators_network_for_time_series_processing}, the task-level RON reservoirs adopt $n_y=13000$ hidden units for sMNIST, $n_y=100$ for ADIAC, $n_y=1000$ for Mackey--Glass, and $n_y=6800$ for Lorenz96. Directly using these dimensions as pretraining references would require soft robot reservoirs with correspondingly many generalized coordinates and repeated differentiable simulations at physically meaningful time steps, making the optimization prohibitively expensive. For this reason, we adopt a simplification by taking four different configurations\footnote{Four configurations are considered instead of just one in order to study the scalability of the approach to bigger reservoirs.} (for each task) with $n_y=6, \;9, 12$ and $15$. This certainly limits the performance  of the model, however, since the aim of this work is showing that an optimization procedure of the physical reservoir can be effective in improving the performance of physical reservoir computing with respect to non-optimized reservoirs, having such a simplification is still sufficient for this purpose.

In total, this yields 16 configurations. Following~\cite{random_oscillators_network_for_time_series_processing}, these configurations were selected by manual tuning to identify a high-performing setup for each scenario. In particular, the discretization step $\Delta t$, spectral radius $\rho_\mathrm{w}$, and input scaling $\nu$ were tuned by exploring values in $\Delta t\in[0.001, \; 0.2]$, $\rho_\mathrm{w}\in[0.9, \; 9]$ and $\nu\in[0.01, \; 10]$; $\vect{W},\vect{V},\vect{b}$ were then generated from those as specified in~\cite{random_oscillators_network_for_time_series_processing}. The damping factors $\bm{\varepsilon}$ and stiffnesses $\bm{\gamma}$ were instead uniformly sampled within the ranges $[\bm{\varepsilon}_\mathrm{min},\, \bm{\varepsilon}_\mathrm{max}]$ and $[\bm{\gamma}_\mathrm{min},\, \bm{\gamma}_\mathrm{max}]$. All specific values used for this selection are reported in Tab.~\ref{tab:RON_datasets}.

\begin{table}[t]
    \small
    \centering
    \setlength{\tabcolsep}{6pt}
    \renewcommand{\arraystretch}{1.0}
    \begin{tabular}{llcccccccc}
        \toprule
        \textbf{Task} & $\bm{n_y}$ & $\bm{\Delta t}$ & $\bm{\rho_\mathrm{w}}$ & $\bm{\nu}$ & $\bm{\gamma_\mathrm{min}}$ & $\bm{\gamma_\mathrm{max}}$ & $\bm{\varepsilon_\mathrm{min}}$ & $\bm{\varepsilon_\mathrm{max}}$ & \textbf{Performance$^*$} \\
        \midrule
        \multirow{4}{*}{\makecell[l]{sMNIST\\($^*$\% $\uparrow$)}}
            &  6 & 0.006 & 0.99 & 1 & 1.7 & 3.7 & 0.01 & 1.01 & 65.88 \\
            &  9 & 0.006 & 0.99 & 1 & 1.7 & 3.7 & 0.01 & 1.01 & 68.66 \\
            & 12 & 0.006 & 0.99 & 1 & 1.7 & 3.7 & 0.01 & 1.01 & 70.51 \\
            & 15 & 0.010 & 9.00 & 1 & 1.7 & 3.7 & 0.01 & 1.01 & 78.57 \\
        \midrule
        \multirow{4}{*}{\makecell[l]{ADIAC\\($^*$\% $\uparrow$)}}
            &  6 & 0.02 & 0.99 & 10 & 2 & 4 & 4.5 & 5.5 & 46.80 \\
            &  9 & 0.02 & 0.99 & 10 & 2 & 4 & 4.5 & 5.5 & 53.20 \\
            & 12 & 0.02 & 9.00 & 10 & 2 & 4 & 4.5 & 5.5 & 57.03 \\
            & 15 & 0.02 & 9.00 & 10 & 2 & 4 & 4.5 & 5.5 & 62.15 \\
        \midrule
        \multirow{4}{*}{\makecell[l]{Mackey--Glass\\($^*$NRMSE $\downarrow$)}}
            &  6 & 0.05 & 0.9 & 1  & 0.1 & 10.1 & 1.5 & 2.5 & 0.513 \\
            &  9 & 0.15 & 0.9 & 10 & 1.0 & 3.0 & 1.5 & 2.5 & 0.432 \\
            & 12 & 0.15 & 0.9 & 10 & 1.0 & 3.0 & 1.5 & 2.5 & 0.358 \\
            & 15 & 0.15 & 3.0 & 10 & 1.0 & 3.0 & 1.5 & 2.5 & 0.309 \\
        \midrule
        \multirow{4}{*}{\makecell[l]{Lorenz96\\($^*$NRMSE $\downarrow$)}}
            &  6 & 0.05 & 0.99 & 0.01 & 9 & 11 & 9.5 & 10.5 & 0.564 \\
            &  9 & 0.05 & 0.99 & 0.01 & 9 & 11 & 9.5 & 10.5 & 0.515 \\
            & 12 & 0.17 & 0.99 & 0.10 & 9 & 11 & 9.5 & 10.5 & 0.417 \\
            & 15 & 0.17 & 0.99 & 0.10 & 9 & 11 & 9.5 & 10.5 & 0.385 \\
        \bottomrule
    \end{tabular}
    \caption{RON parameter choices for the reference dynamics used in the optimization. For each task, 4 datasets are collected with 4 different reservoir dimensions $n_y$. The parameters $\Delta t$, $\nu$ and the spectral radius $\rho_\mathrm{w}$ of $\vect{W}$ are selected through manual tuning, while $\vect{W}$, $\vect{V}$, $\vect{b}$ are generated as specified in~\cite{random_oscillators_network_for_time_series_processing}; $\bm{\varepsilon}$ and $\bm{\gamma}$ are instead uniformly sampled within ranges $[\bm{\varepsilon}_\mathrm{min},\, \bm{\varepsilon}_\mathrm{max}]$ and $[\bm{\gamma}_\mathrm{min},\, \bm{\gamma}_\mathrm{max}]$. Performance is evaluated as \% classification accuracy ($\uparrow$) for sMNIST and ADIAC, and as prediction NRMSE ($\downarrow$) for Mackey--Glass and Lorenz96.}
    \label{tab:RON_datasets}
\end{table}

\paragraph{Pretraining dataset generation} After defining the required configurations, 16 datasets (one for each) were collected for the physical reservoir optimization. For each dataset, $m=10^5$ data points $(\vect{y},\dot{\vect y},\vect{u})$ were sampled, and the corresponding labels $\ddot{\vect y}$ were computed using Eq.~\eqref{eq:RON_equation}. The data points were sampled uniformly within ranges determined by simulating the RON with a random input sequence consistent with the task at hand and observing its response $\vect{y}(t),\dot{\vect{y}}(t)$. The RON model and related components were implemented using a modified version of the publicly available Archetype Computing and Adaptive System (ACDS) library~\cite{acds_library}, which is based on PyTorch~\cite{pytorch}.

\section{Planar Piecewise Constant Strain Soft Robot Model}
\label{app:pcs_model}
In this work, the soft robots in the physical reservoir are modeled as slender compliant arms and represented using the planar Piecewise Constant Strain (PCS) model~\cite{pcs_paper1,pcs_paper2}. This appendix briefly introduces the model and the assumptions adopted for its implementation.

The PCS is a kinematic model based on a finite-dimensional parametrization $\vect{q}$ that describes the robot as a backbone curve composed of $n_\mathrm{pcs}$ segments. Each $i$-th segment has constant (in space) strains (bending $\K^{(i)}$ [rad/m], axial $\sigma_\mathrm{ax}^{(i)}$ [--] and shear $\sigma_\mathrm{sh}^{(i)}$ [--]) and its own physical properties: length $L^{(i)}$ [m], damping matrix $\vect{D}^{(i)}=\mathrm{diag}([d^{(i)}_\mathrm{be},\;d^{(i)}_\mathrm{ax},\;d^{(i)}_\mathrm{sh}])$ [Pa$\cdot$s], radius $r^{(i)}$ [m], density $\rho^{(i)}$ [kg/m$^3$], elastic modulus $E^{(i)}$ [Pa], and shear modulus $G^{(i)}$ [Pa], with $i=1, \, ... \, ,n_\mathrm{pcs}$. The kinematics of the soft robot can then be expressed as a function of the generalized coordinates $\vect{q}=[\K^{(1)},\;\sigma_\mathrm{ax}^{(1)},\;\sigma^{(1)}_\mathrm{sh},\; ... \ ,\;\K^{(n_\mathrm{pcs})},\;\sigma_\mathrm{ax}^{(n_\mathrm{pcs})},\;\sigma^{(n_\mathrm{pcs})}_\mathrm{sh}]^\top\in\R^{3 n_\mathrm{pcs}}$, while its dynamics are obtained by applying the Euler-Lagrange approach. The definitions of the dynamical matrices and kinematic equations are provided in~\cite{pcs_paper2}.

\begin{table}[!b]
    \small
    \centering
    \setlength{\tabcolsep}{6pt}
    \renewcommand{\arraystretch}{1.0}
    \begin{tabular}{lcccccc}
        \toprule
        \multirow{2}{*}{\textbf{Config.}} & \multirow{2}{*}{\textbf{Run n.}}
        & \multicolumn{2}{c}{\textbf{Performance (task NRMSE [-])}} & \multicolumn{2}{c}{\textbf{Elapsed time (simulation) [s]}} & \multirow{2}{*}{\makecell{\textbf{Strain trajectories} \\ \textbf{NRMSE($\vect{q}_\mathrm{c},\vect{q}_\mathrm{nc}$) [-]}}} \\
        \cmidrule(lr){3-4} \cmidrule(lr){5-6}
        & & \makecell{No Coriolis \\ and centrifugal} & \makecell{With Coriolis \\ and centrifugal} & \makecell{No Coriolis \\ and centrifugal} & \makecell{With Coriolis \\ and centrifugal} \\
        \midrule
        \multirow{4}{*}{$n_y=6$}
            & 1 & 0.5300 & 0.5456 & 12.2 & 17.1 & $1.10\cdot10^{-3}$ \\
            & 2 & 0.5230 & 0.5232 & 12.4 & 17.5 & $1.49\cdot10^{-5}$ \\
            & 3 & 0.5236 & 0.5415 & 12.0 & 184.0$^*$ & $2.44\cdot10^{-4}$\\
            & 4 & 0.5241 & 0.5235 & 12.7 & 18.4 & $3.97\cdot10^{-5}$ \\
        \midrule
        \multirow{4}{*}{$n_y=9$}
            & 1 & 0.4982 & 0.5058 & 36.9 & 488.3$^*$ & $7.84\cdot10^{-5}$ \\
            & 2 & 0.5000 & 0.5012 & 37.3 & 51.9 & $1.18\cdot10^{-5}$ \\
            & 3 & 0.5296 & 0.5333 & 38.9 & 50.5 & $1.74\cdot10^{-6}$ \\
            & 4 & 0.4930 & 0.4966 & 38.0 & 51.4 & $4.24\cdot10^{-4}$ \\
        \bottomrule
    \end{tabular}
    \caption{Representative tests studying the influence of Coriolis and centrifugal forces on the system. These are the results of PRC benchmark experiments for the Mackey--Glass task, with reservoir dimension $n_y\in\{6,9\}$; all 4 runs with different seeds are reported. The physical reservoir that was previously pretrained (neglecting Coriolis and centrifugal effects in the soft robots) is simulated (both accounting and not accounting for these forces) and employed to train and test the output layer. The test NRMSE on the Mackey--Glass task and elapsed simulation time during testing are reported above. The NRMSE between robot's strains with and without Coriolis and centrifugal terms is reported as well. An asterisk denotes that it was necessary to reduce the simulation step from $\delta t=0.1$ ms to $\delta t=0.01$ ms in order to obtain a consistent simulation.}
    \label{tab:coriolis}
\end{table}

\begin{figure}[t]
    \centering
    \includegraphics[width=1\textwidth]{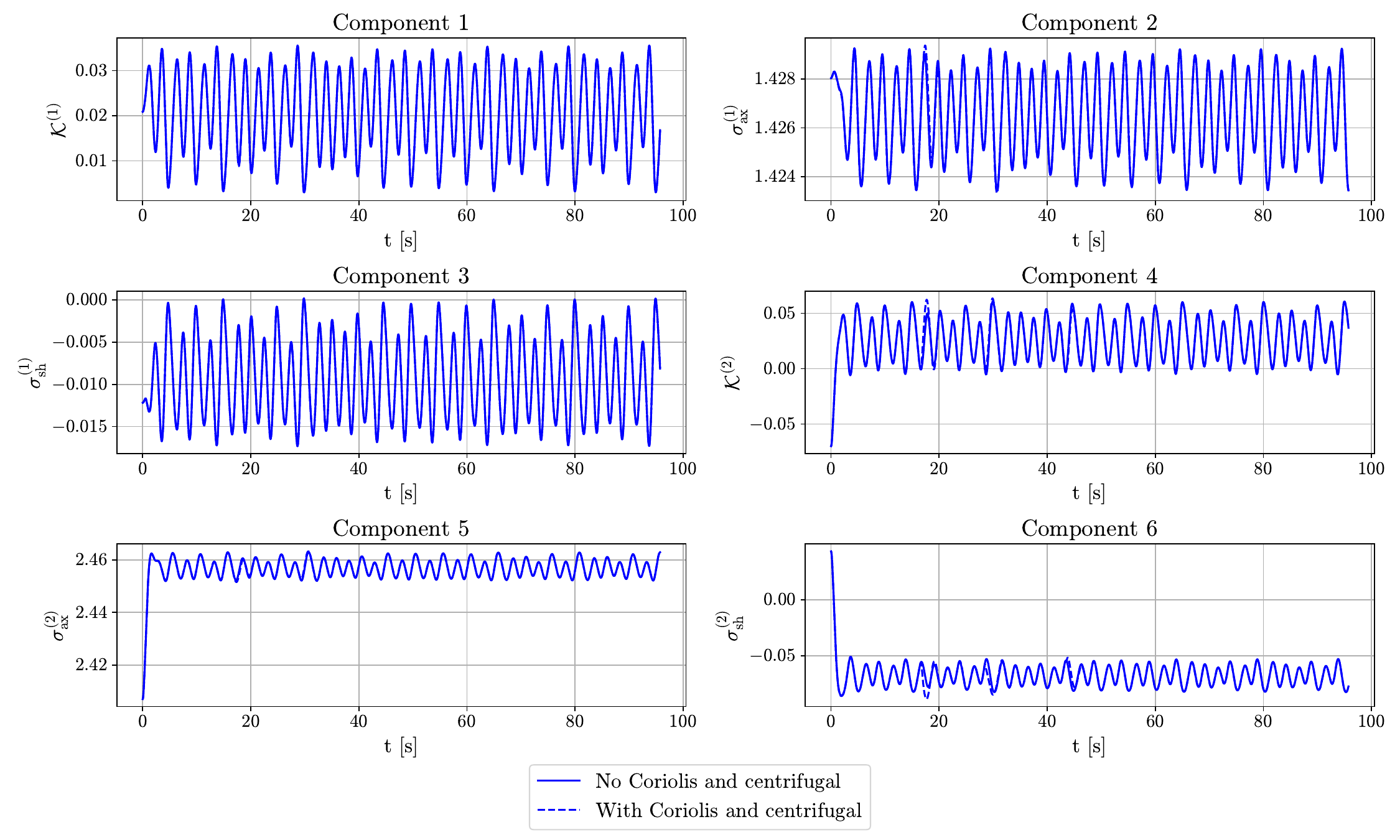}
    \caption{Soft robot strains in the physical reservoir during inference in the Mackey--Glass task with $n_y=6$. The plot shows both the simulation that accounts for the Coriolis and centrifugal effects (dashed line) and that which neglects them (solid line).}
    \label{fig:coriolis_plot}
\end{figure}

The soft robot implementation uses the SoRoMoX library~\cite{soromox_paper, soromox_package} in Python and JAX~\cite{jax2018github}, which supports JIT compilation and is optimized for automatic differentiation and parallelization. For each robot, actuation was modeled directly in generalized-coordinate space as $\bm{\tau}(t)=[\tau_\mathrm{be}^{(1)},\;\tau^{(1)}_\mathrm{ax},\;\tau^{(1)}_\mathrm{sh},\;... \; , \; \tau_\mathrm{be}^{(n_\mathrm{pcs})},\;\tau^{(n_\mathrm{pcs})}_\mathrm{ax},\;\tau^{(n_\mathrm{pcs})}_\mathrm{sh}]^\top\in\R^{3n_\mathrm{pcs}}$, where $\tau_\mathrm{be}^{(i)}$ [N$\cdot$m$^2$] and $\tau_\mathrm{ax}^{(i)},\tau_\mathrm{sh}^{(i)}$ [N$\cdot$m] denote the generalized bending, axial, and shear inputs associated with segment $i$. These inputs correspond to the generalized forces appearing in the virtual-work expression of a fully actuated model. The strain coordinates $\vect{q}(t)$ were assumed to be measurable and were used as reservoir outputs.

Finally, centrifugal and Coriolis effects were neglected because their evaluation is computationally expensive, often requiring a smaller $\delta t$ to ensure consistency in the simulation, while their inclusion was observed to have a negligible impact on the rollout behavior of the soft robot. In fact, Tab.~\ref{tab:coriolis} displays the Normalized Root Mean Square Error (NRMSE)
\begin{equation*}
    \mathrm{NRMSE}(\vect{q}_\mathrm{c},\vect{q}_\mathrm{nc})=\frac{\sqrt{\sum_{k=0}^{n_\mathrm{steps}}\left\|\vect{q}_\mathrm{c}(k) - \vect{q}_\mathrm{nc}(k)\right\|_2^2}}{\sqrt{\sum_{k=0}^{n_\mathrm{steps}}\left\|\vect{q}_\mathrm{nc}(k)\right\|_2^2}}
\end{equation*}
between robot's strains histories with ($\vect{q}_\mathrm{c}$) and without ($\vect{q}_\mathrm{nc}$) Coriolis and centrifugal terms for some representative experiments, and these values are very low in all cases: Fig.~\ref{fig:coriolis_plot} gives a visual comparison for one specific case. Moreover, Tab.~\ref{tab:coriolis} also shows that accounting for Coriolis and centrifugal forces changes the PRC performance---in terms of test NRMSE on the Mackey--Glass task---only by \SI{1.63}{\percent} (for $n_y=6$) and \SI{0.80}{\percent} (for $n_y=9$) on average across the four runs. On the other side, the computation time for simulating the system and generating the activations for the output layer increases significantly, especially when it is required to reduce the simulation step $\delta t$, showing an average growth of \SI{42.1}{\percent} (for $n_y=6$) and \SI{34.7}{\percent} (for $n_y=9$) without even accounting for such cases. Notice that Tab.~\ref{tab:coriolis} only presents some representative tests, with Mackey--Glass task and $n_y\in\{6,9\}$, but the influence of Coriolis and centrifugal effects on the computational time is even more critical for higher reservoir dimensions ($n_y\in\{12,15\}$) and for the other tasks, since they require more and/or longer simulation rollouts.

\section{Physical Reservoir Optimization Protocol}
\label{app:optimization_algorithm}
This appendix presents hardware, software, algorithms and technical details concerning the implementation of the co-optimization process of the physical reservoir.

Most optimization runs were performed on a desktop workstation equipped with a 24-core Intel Core i9-12900KF processor, \SI{64}{GB} of RAM, and an NVIDIA GeForce RTX 3090 GPU with \SI{24}{GB} of VRAM, running Ubuntu 22.04. The implementation relies on Python 3.11, JAX~\cite{jax2018github} and jaxlib 0.8.1 for efficient automatic differentiation, Optax 0.2.6~\cite{optax_deepmind2020jax}, Diffrax 0.7.0~\cite{diffrax} and SoRoMoX~\cite{soromox_paper, soromox_package}. 

Concerning the initialization of the optimization variables, $\theta_\mathrm{contr}$ in the MLP controllers were initialized according to the Glorot/Xavier approach~\cite{glorot_xavier_initialization}. The mapping parameters $\theta_\mathrm{map}$ were initialized as $\vect{c}=\vect{0}$ and $\vect{A}=\vect{U}\bm{\Sigma}\vect{V}^\top$, with $\vect{U}$ and $\vect{V}$ orthogonal matrices sampled from a normal distribution and $\bm{\Sigma}=\mathrm{diag}([0.003,0.003,\;...])$. The robot parameters $\theta_\mathrm{rob}$ were sampled from a uniform distribution with the ranges reported in Tab.~\ref{tab:robot_initialization}.

The computation time for the co-optimization process depends on many factors, such as the dimension of the reservoir and the number of optimization parameters. In general, a single pretraining run was observed to last indicatively between 50 and 90 minutes.

\begin{table}[!t]
    \small
    \centering
    \setlength{\tabcolsep}{5pt}
    \renewcommand{\arraystretch}{1.0}
    \begin{tabular}{lcccc}
        \toprule
        \textbf{Parameter} & \textbf{Symbol} & \textbf{Unit} & \textbf{Min value} & \textbf{Max value} \\
        \midrule
        length & $L$ & [m] & 0.01 & 0.80 \\
        damping (bending) & $d_\mathrm{be}$ & [Pa s] & $5\cdot 10^{-7}$ & $5\cdot 10^{-5}$ \\
        damping (axial) & $d_\mathrm{ax}$ & [Pa s] & $5\cdot 10^{-4}$ & $5\cdot 10^{-2}$ \\
        damping (shear) & $d_\mathrm{sh}$ & [Pa s] & $5\cdot 10^{-4}$ & $5\cdot 10^{-2}$ \\
        radius & $r$ & [m] & 0.005 & 0.070 \\
        density & $\rho$ & [kg/m$^3$] & 900 & 1200 \\
        Young modulus & $E$ & [Pa] & 1500 & 2500 \\
        shear modulus & $G$ & [Pa] & 700 & 1300 \\
        \bottomrule
    \end{tabular}
    \caption{Sampling ranges for the initialization of the soft robot optimization variables. Each value is referred to the single $i$-th segment in the PCS model.}
    \label{tab:robot_initialization}
\end{table}
The co-optimization problem was solved using parallel multi-start gradient descent with Adam~\cite{adam_a_method_for_stochastic_optimization} updates. Each start was run for $1500$ epochs with gradient clipping and a cosine-decay learning rate from $10^{-3}$ to $10^{-6}$, preceded by a linear warm-up of $100$ epochs. These hyperparameters were manually tuned to achieve a reasonable computation time while maintaining a final validation loss low enough to enable an accurate reconstruction of the nominal dynamics. The reservoir pretraining dataset was split into \SI{70}{\percent} training data, \SI{10}{\percent} validation data, and \SI{20}{\percent} test data.

\section{Statistical Significance of the Benchmark Results}
\label{app:mann_whitney_test}

\begin{table}[!b]
    \small
    \centering
    \setlength{\tabcolsep}{6pt}
    \renewcommand{\arraystretch}{1.0}
    \begin{minipage}{0.47\textwidth}
        \centering
        \begin{tabular}{llcc}
            \toprule
            \textbf{Task} & $\bm{n_y}$ & \makecell{$\bm{p}$\textbf{-value} \\ \textbf{vs \textit{unoptimized}}} & \makecell{$\bm{p}$\textbf{-value} \\ \textbf{vs \textit{partially}}\\ \textbf{\textit{pretrained}}} \\
            \midrule
                \multirow{4}{*}{sMNIST} & 6 & \textbf{0.0143} & \textbf{0.0143} \\
                 & 9 & \textbf{0.0143} & \textbf{0.0143} \\
                 & 12 & \textbf{0.0143} & 0.0571 \\
                 & 15 & \textbf{0.0143} & 0.1714 \\
            \midrule
                \multirow{4}{*}{ADIAC} & 6 & \textbf{0.0143} & 0.1000 \\
                 & 9 & \textbf{0.0143} & 0.0571 \\
                 & 12 & \textbf{0.0143} & \textbf{0.0143} \\
                 & 15 & 0.2429 & \textbf{0.0143} \\
            \bottomrule
        \end{tabular}
    \end{minipage}
    \hfill
    \begin{minipage}{0.52\textwidth}
        \centering
        \begin{tabular}{llcc}
            \toprule
            \textbf{Task} & $\bm{n_y}$ & \makecell{$\bm{p}$\textbf{-value} \\ \textbf{vs \textit{unoptimized}}} & \makecell{$\bm{p}$\textbf{-value} \\ \textbf{vs \textit{partially}}\\ \textbf{\textit{pretrained}}} \\
            \midrule
                \multirow{4}{*}{Mackey--Glass} & 6 & \textbf{0.0143} & 0.1714 \\
                 & 9 & \textbf{0.0143} & \textbf{0.0286} \\
                 & 12 & \textbf{0.0143} & \textbf{0.0286}$^*$ \\
                 & 15 & \textbf{0.0143} & \textbf{0.0143} \\
            \midrule
                \multirow{4}{*}{Lorenz96} & 6 & \textbf{0.0143} & \textbf{0.0143} \\
                 & 9 & \textbf{0.0143} & \textbf{0.0143} \\
                 & 12 & \textbf{0.0286} & \textbf{0.0143} \\
                 & 15 & \textbf{0.0286} & \textbf{0.0286} \\
            \bottomrule
        \end{tabular}
    \end{minipage}
    \caption{One-sided Mann--Whitney $U$ test $p$-values for the alternative hypothesis that the fully pretrained physical reservoir (\textit{ours}) outperforms each baseline, computed from the per-seed benchmark results (four runs per configuration). Bold values indicate statistical significance at the $\alpha=0.05$ level ($p<0.05$) and thus rejection of the null hypothesis. Diverged trials are retained in the rank-based analysis; an asterisk denotes a configuration in which a NaN run reduced the baseline sample size to three.}
    \label{tab:mann_whitney_test}
\end{table}

To assess whether the improvements of the fully pretrained reservoir over the other baselines are statistically meaningful rather than an artifact of the particular random seeds, we compare the per-seed benchmark results with a one-sided Mann--Whitney \textit{U} test~\cite{mann_whitney_u_test}. This non-parametric, rank-based test is well-suited for the small number of samples considered here (four runs per configuration), as it makes no assumptions about the distributions and it is robust to the occasional diverged trial, which only enters as the worst rank. For each task and reservoir dimension $n_y$, we test the alternative hypothesis that our methodology outperforms a given baseline, using the exact null distribution. We adopt a significance level of 0.05, meaning that a $p$-value below 0.05 allows us to reject the null hypothesis that the two sets of per-seed results are drawn from the same distribution. NaN runs are excluded, reducing the corresponding samples to 3, while diverged runs are retained, since the rank-based test is insensitive to their magnitude.

The resulting $p$-values are reported in Tab.~\ref{tab:mann_whitney_test}. They provide sufficient evidence that the fully pretrained reservoir significantly outperforms the \textit{unoptimized} baseline ($p<0.05$) in 15 out of the 16 configurations, and the stronger \textit{partially pretrained} baseline in 11 of the 16. The six non-significant cases almost all involve the partially pretrained baseline at settings where the two distributions partially overlap.

Overall, these tests confirm that the improvement of the fully pretrained reservoir is statistically significant in the large majority of the considered settings (26 of the 32 comparisons). We note that, with four runs per configuration, the smallest attainable one-sided $p$-value is $1/\binom{8}{4}\approx0.0143$, which is why the significant entries in Tab.~\ref{tab:mann_whitney_test} often assume this specific value. Therefore, the test has limited resolution at this sample size, and the reported $p$-values should be read together with the consistently low standard deviations of the \textit{pretrained (ours)} approach---those observed during the benchmarking experiments---as complementary evidence of a robust improvement.

\section{Digital Inference Cost of Digital and Physical Reservoir Pipelines}
\label{app:flops_count}
Physical Reservoir Computing (PRC) can reduce digital inference cost because the reservoir dynamics evolve directly in a physical substrate rather than being computed through numerical simulation. The proposed architecture nevertheless retains digital feedforward and feedback controllers, together with an affine state map and a task-specific readout. We therefore quantify whether the online cost of these digital components preserves the computational advantage of the physical reservoir relative to the digital Random Oscillators Network (RON).

We compare the two inference pipelines in terms of floating-point operations (FLOPs). For the PRC pipeline, the soft robot dynamics are assumed to evolve physically and therefore incur no digital FLOPs. Accordingly, this analysis does not quantify sensing, actuation, communication, physical energy consumption, or wall-clock latency. We count the evaluation of $\tanh(\vect{x})$ or $\mathrm{softmax}(\vect{x})$ as $n$ FLOPs for $\vect{x}\in\R^n$. Here, $n_y$ is the reservoir dimension, $n_u$ is the input dimension, $K$ is the number of input time steps, $n_c$ is the number of classes in classification tasks, and $N_l$ is the prediction lag in forecasting tasks; the subscripts $_\mathrm{cls}$ and $_\mathrm{for}$ denote classification and forecasting, respectively.

The affine physical-to-reference map can be absorbed into the readout, thereby avoiding a separate online mapping operation. Before this substitution, the task-specific readouts operate on the scaled reservoir variables $\vect{\tilde y}$ as
\begin{align}
    \vect{r}_{\mathrm{cls}}
    &=\mathrm{softmax}\!\left(\vect{W}_o \; \vect{\tilde y}(K)+\vect{b}_o\right)
      \in\R^{n_c},
    \label{eq:readout_classification}\\
    \vect{r}_{\mathrm{for}}(k+N_l)
    &=\vect{W}_o \; \vect{\tilde y}(k)+\vect{b}_o
      \in\R^{n_u}.
    \label{eq:readout_forecasting}
\end{align}
Here, $\vect{W}_o$ and $\vect{b}_o$ denote the trained readout parameters with the appropriate output dimension for each task (Appendix~\ref{app:tasks}). To express these readouts directly in terms of the physical reservoir coordinates, let
\begin{equation*}
    \vect{D}_{\sigma}=\operatorname{diag}(\bm{\sigma})^{-1}\in\R^{n_y \times n_y},
\end{equation*}
where $\bm{\mu}$ and $\bm{\sigma}$ are the mean and standard deviation used to scale the reference states (Appendix~\ref{app:tasks}). Substituting the inverse affine map $\vect{y}(k)=\vect{A}^{-1}(\vect{Q}(k)-\vect{c})$ into the scaling operation gives
\begin{align*}
    \tilde{\vect{y}}(k)
    &=\vect{D}_{\sigma}\bigl(\vect{y}(k)-\bm{\mu}\bigr)\\
    &=\vect{D}_{\sigma}
      \left[\vect{A}^{-1}\bigl(\vect{Q}(k)-\vect{c}\bigr)-\bm{\mu}\right]\\
    &=\vect{D}_{\sigma}\vect{A}^{-1}\vect{Q}(k)
      -\vect{D}_{\sigma}\left(\vect{A}^{-1}\vect{c}+\bm{\mu}\right).
\end{align*}
The derivation first replaces the scaled reservoir variables with its definition, then substitutes the inverse map, expands the affine terms, and finally collects the factors that depend on $\vect{Q}(k)$ and those that remain constant. Applying the trained readout parameters $\vect{W}_o$ and $\vect{b}_o$ therefore yields the task-specific coefficients
\begin{equation*}
    \vect{B}=\vect{W}_o\vect{D}_{\sigma}\vect{A}^{-1},
    \qquad
    \vect{C}=\vect{W}_o\vect{D}_{\sigma}
    \left(\vect{A}^{-1}\vect{c}+\bm{\mu}\right)-\vect{b}_o.
\end{equation*}
For a readout dimension $d$, then $\vect{B}\in\R^{d\times n_y}$ is the effective matrix that maps the physical reservoir coordinates $\vect{Q}$ directly to the readout logits or forecasts. The vector $\vect{C}\in\R^d$ is the corresponding effective offset, which combines the affine-map translation, the scaling mean, and the trained readout bias. Here, $d=n_c$ for classification and $d=n_u$ for forecasting. The two readouts in Eqs.~\eqref{eq:readout_classification} and \eqref{eq:readout_forecasting} can thus be written compactly as
\begin{align*}
    \vect{r}_{\mathrm{cls}}
    &=\mathrm{softmax}\!\left(\vect{B}\vect{Q}(K)-\vect{C}\right)
      \in\R^{n_c},\\
    \vect{r}_{\mathrm{for}}(k+N_l)
    &=\vect{B}\vect{Q}(k)-\vect{C}
      \in\R^{n_u},
\end{align*}
for classification and forecasting, respectively.
In the forecasting case, the physical reservoir state at time $k$ predicts the target at time $k+N_l$. The task-specific coefficients $\vect{B}$ and $\vect{C}$ are computed once after readout training, so only the resulting task-specific readout is evaluated online.

\subsubsection{Digital cost of reservoir dynamics}
\begin{itemize}
    \item The RON evaluates the discrete state update in Eq.~\eqref{eq:RON_discrete}, requiring
    \begin{equation*}
        \mathrm{FLOPs}=K \; (2n_y^2+9n_y+2n_y n_u).
    \end{equation*}
    \item Within the reservoir component of the proposed PRC system, only the feedforward and feedback controllers are evaluated digitally. For the controller architecture used in this work, their cost is
    \begin{equation}
        \mathrm{FLOPs}=K \; (385n_y+128 n_u+16640).
        \label{eq:flops_phys_reservoir}
    \end{equation}
\end{itemize}

\subsubsection{Digital cost of the task-specific readout}
After incorporating the affine state map into the coefficients $\vect{B}$ and $\vect{C}$, the RON and PRC pipelines have the same online readout cost:
\begin{itemize}
    \item For classification,
    \begin{equation*}
        \mathrm{FLOPs}=2n_y n_c+3 n_c.
    \end{equation*}
    \item For forecasting,
    \begin{equation*}
        \mathrm{FLOPs}=(K-N_l) \; 2n_y n_u.
    \end{equation*}
\end{itemize}

\subsubsection{Digital cost of the complete inference pipeline}
\begin{figure}[!t]
    \centering
    \includegraphics[width=0.8\textwidth]{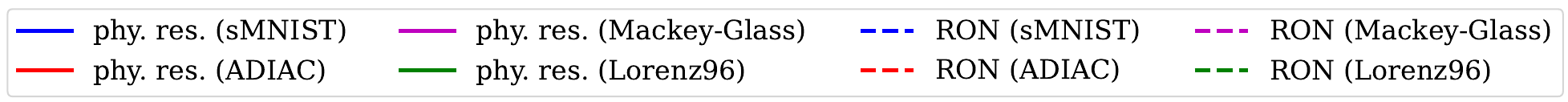}
    \\[0.5em]
    \begin{subfigure}[b]{0.32\textwidth}
        \centering
        \includegraphics[width=\textwidth]{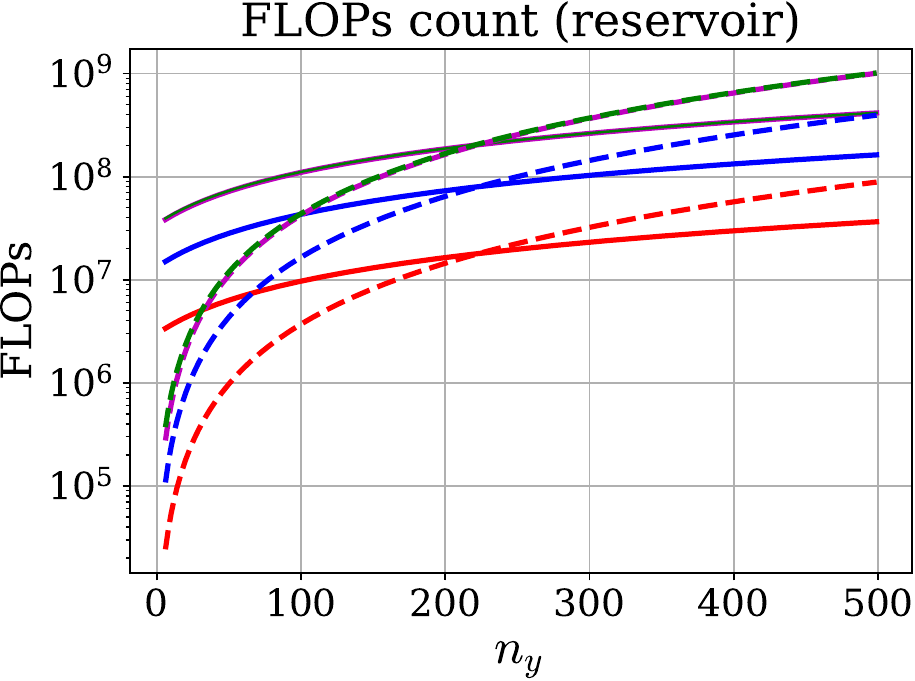}
        \caption{Digital FLOPs required by the reservoir dynamics.}
        \label{fig:flops_reservoir}
    \end{subfigure}
    \hfill
    \begin{subfigure}[b]{0.32\textwidth}
        \centering
        \includegraphics[width=\textwidth]{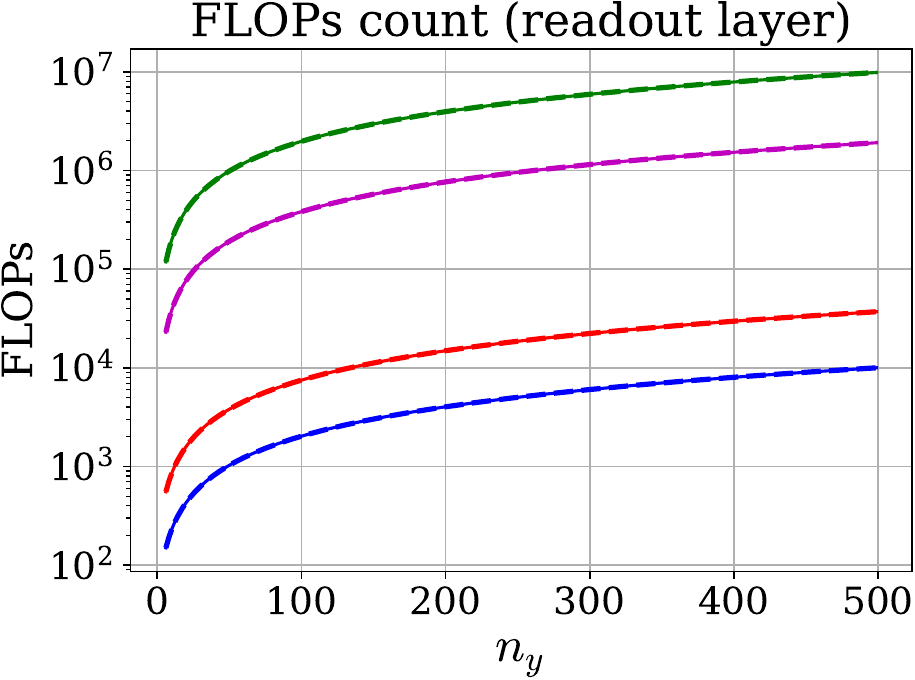}
        \caption{Digital FLOPs required by the task-specific readout.}
        \label{fig:flops_readout}
    \end{subfigure}
    \hfill
    \begin{subfigure}[b]{0.32\textwidth}
        \centering
        \includegraphics[width=\textwidth]{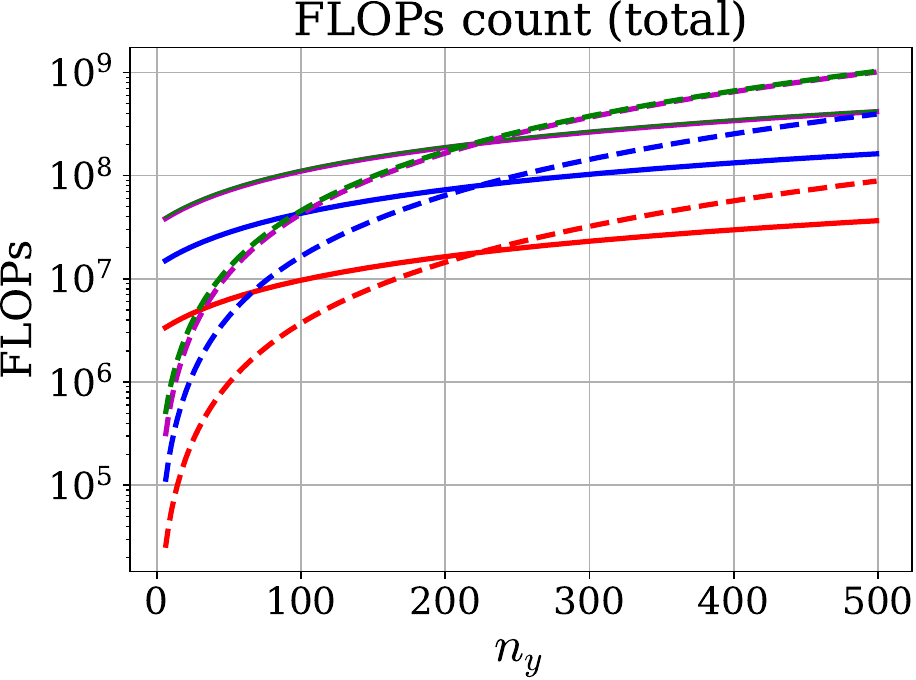}
        \caption{Digital FLOPs required by the complete inference pipeline.}
        \label{fig:flops_tot}
    \end{subfigure}
    \caption{Digital inference cost comparison for the four benchmark tasks, separated into reservoir dynamics, task-specific readout, and complete pipeline costs. Input dimensions, time-series lengths, and other task-specific parameters are those adopted in our work.}
    \label{fig:flops}
\end{figure}

Combining the reservoir and readout contributions gives the following total digital costs:
\begin{itemize}
    \item For the RON,
    \begin{equation*}
        \mathrm{FLOPs}=K \; (2n_y^2+9n_y+2n_y n_u) + 2n_y n_c+3 n_c \quad \text{for classification};
    \end{equation*}
    \begin{equation*}
        \mathrm{FLOPs}=K \; (2n_y^2+9n_y+2n_y n_u) + (K-N_l) \; 2n_y n_u \quad \text{for forecasting}.
    \end{equation*}
    \item For the proposed PRC system,
    \begin{equation}
        \mathrm{FLOPs}=K \; (385n_y+128 n_u+16640) + 2n_y n_c+3 n_c \quad \text{for classification};
        \label{eq:flops_prc_classification}
    \end{equation}
    \begin{equation}
        \mathrm{FLOPs}=K \; (385n_y+128 n_u+16640) + (K-N_l) \; 2n_y n_u \quad \text{for forecasting}.
        \label{eq:flops_prc_forecasting}
    \end{equation}
\end{itemize}
For $n_y\gg n_u,n_c$, the total digital cost of the RON scales as $\mathcal{O}(K n_y^2)$ for both task types. The PRC cost instead scales as $\mathcal{O}(K n_y+n_y n_c)$ for classification and $\mathcal{O}(K n_y+(K-N_l)n_y n_u)$ for forecasting. Fig.~\ref{fig:flops} reports the exact FLOP counts for the four benchmark tasks as functions of $n_y$. Because the affine map is incorporated into the readout coefficients, the readout cost is identical for the RON and PRC pipelines. Equating the reservoir-only costs shows that the PRC controllers become less expensive than the digital RON dynamics at approximately $n_y\approx220$ for the benchmark input dimensions.

\begin{figure}[!b]
    \centering
    \includegraphics[width=0.8\textwidth]{Images/legend_flops.pdf}
    \\[0.5em]
    \begin{subfigure}[b]{0.32\textwidth}
        \centering
        \includegraphics[width=\textwidth]{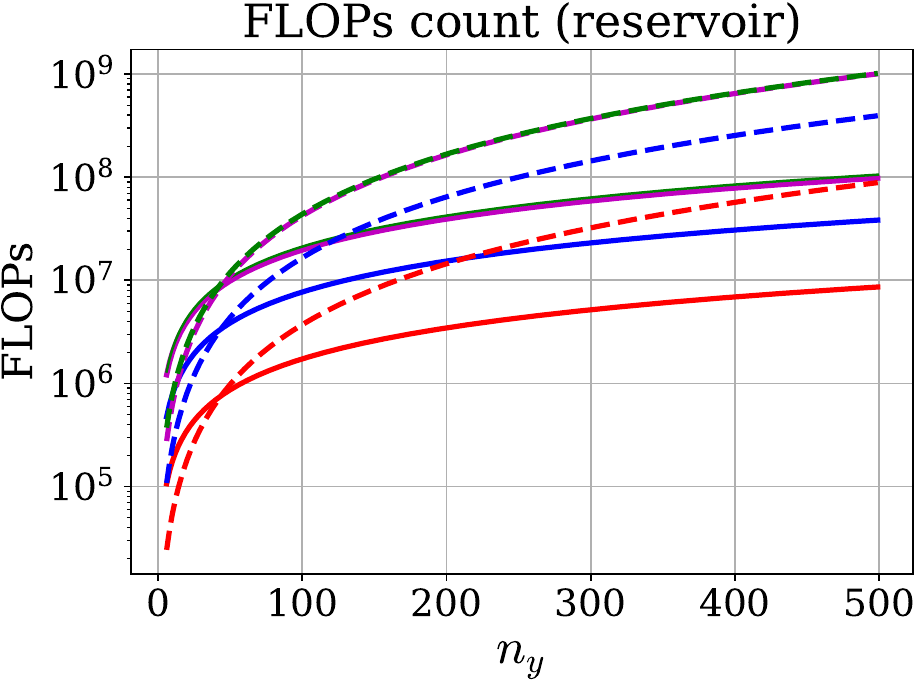}
        \caption{Digital FLOPs required by the reservoir dynamics.}
        \label{fig:flops_reservoir_new}
    \end{subfigure}
    \hfill
    \begin{subfigure}[b]{0.32\textwidth}
        \centering
        \includegraphics[width=\textwidth]{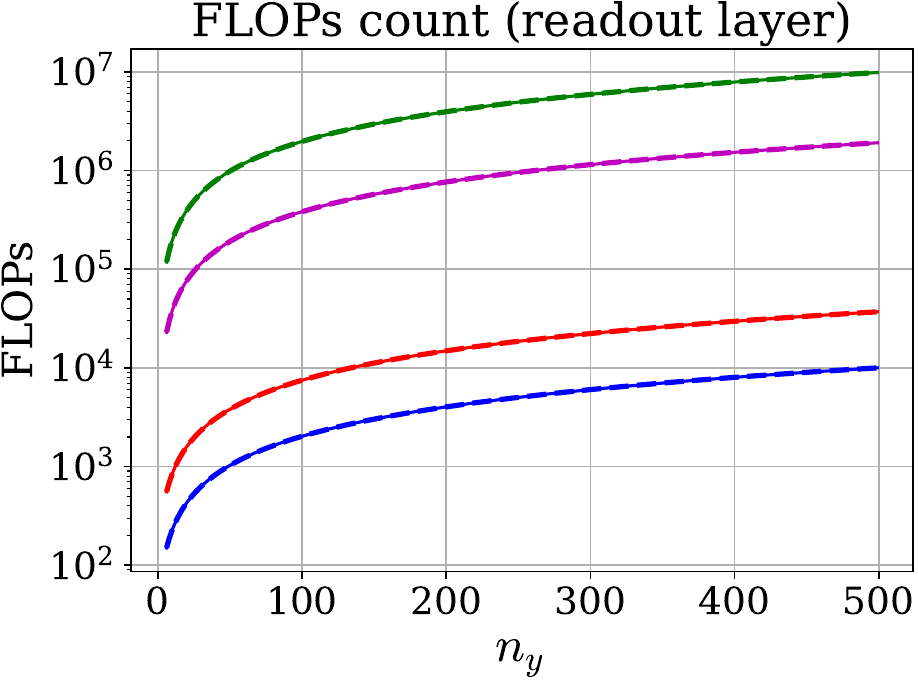}
        \caption{Digital FLOPs required by the task-specific readout.}
        \label{fig:flops_readout_new}
    \end{subfigure}
    \hfill
    \begin{subfigure}[b]{0.32\textwidth}
        \centering
        \includegraphics[width=\textwidth]{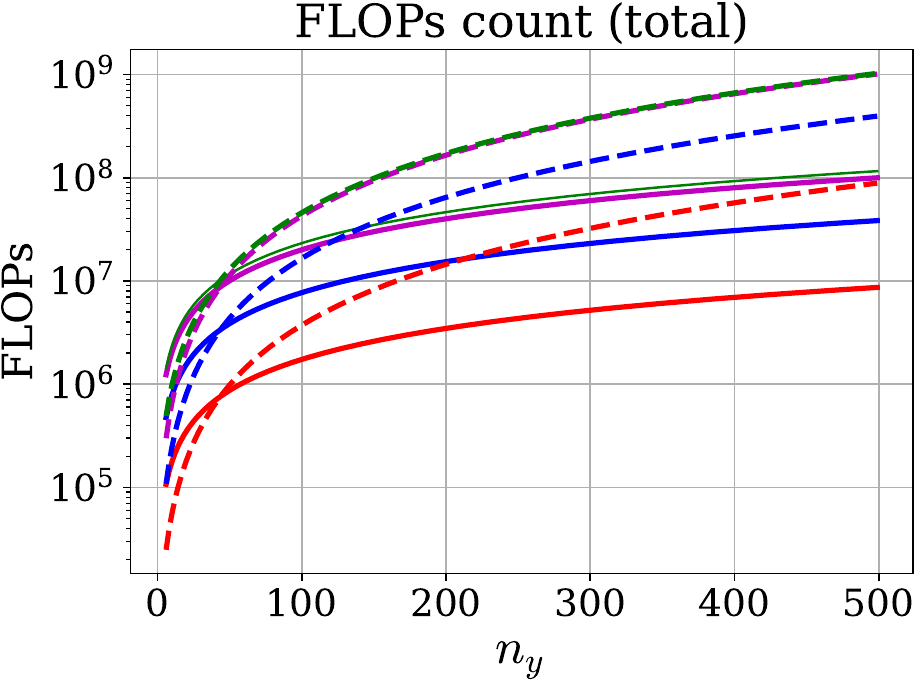}
        \caption{Digital FLOPs required by the complete inference pipeline.}
        \label{fig:flops_tot_new}
    \end{subfigure}
    \caption{Digital inference cost with the simplified feedback and feedforward controllers. Comparison for the four benchmark tasks, separated into reservoir dynamics, task-specific readout, and complete pipeline costs.}
    \label{fig:flops_new}
\end{figure}

\subsection{Digital cost for a simplified controller parametrization}

The feedforward and feedback controllers used throughout this work employ a relatively large MLP architecture (2 layers of 64 neurons each) that is fixed across all benchmark tasks to ensure consistency among experiments. Because this architecture was not optimized for digital inference cost, the PRC pipeline requires more FLOPs than the RON at small reservoir dimensions $n_y$. Reducing the MLP size could lower this fixed overhead and extend the range of $n_y$ for which the PRC pipeline is computationally less expensive. In this regard, we performed additional experiments using an MLP feedback controller with one layer of 16 neurons and a linear feedforward controller $\bm{\tau}_\mathrm{ff}(t)=\vect{V}\vect{u}(t)+\vect{d}$. This simpler architecture gives the physical reservoir a FLOPs count of 
\begin{equation*}
    \mathrm{FLOPs}=K \; (96n_y+2n_y n_u+17),
\end{equation*}
which is lower compared to \eqref{eq:flops_phys_reservoir}, considering that $n_u$ is always very small. As a consequence, the whole PRC pipeline requires
\begin{equation*}
    \mathrm{FLOPs}=K \; (96n_y+2n_y n_u+17) + 2n_y n_c+3 n_c \quad \text{for classification},
\end{equation*}
\begin{equation*}
    \mathrm{FLOPs}=K \; (96n_y+2n_y n_u+17) + (K-N_l) \; 2n_y n_u \quad \text{for forecasting},
\end{equation*}
which are once again lower than those in \eqref{eq:flops_prc_classification} and \eqref{eq:flops_prc_forecasting}. For the tasks considered in this work, a comparison between the FLOPs of this new PRC setup and the digital RON is provided in Fig.~\ref{fig:flops_new}: relative to the ``original'' controllers adopted in our work (Fig.~\ref{fig:flops}), now the PRC controllers become less expensive than the digital RON dynamics at approximately $n_y \approx 40$ rather than $n_y \approx 220$.

\begin{table}[!b]
    \centering
    \setlength{\tabcolsep}{6pt}
    \renewcommand{\arraystretch}{1.0}
    \begin{tabular}{llcccc}
        \toprule
        \multirow{2}{*}{\makecell[l]{\textbf{Controller}\\ \textbf{configuration}}} & \multirow{2}{*}{$\bm{n_y}$} & \multicolumn{2}{c}{\textbf{Pretraining}} & \multirow{2}{*}{\makecell{\textbf{PRC performance}\\ \textbf{(accuracy \%)}}} & \multirow{2}{*}{\textbf{FLOPs}} \\
        \cmidrule(lr){3-4}
        & & \textbf{Test RMSE} & \textbf{Trajectory NRMSE} & & \\
        \midrule
        \multirow{4}{*}{\makecell[l]{\textit{``Original'' (in our work)} \\ FB: 64 neurons, 2 layers\\ FF: 64 neurons, 2 layers}}
         & $6$  & $0.072_{0.005}$ & $0.010_{0.003}$ & $64.03_{1.32}$ & $1.50\cdot10^7$ \\
         & $9$  & $0.079_{0.020}$ & $0.010_{0.003}$ & $66.78_{1.82}$ & $1.59\cdot10^7$ \\
         & $12$ & $0.132_{0.004}$ & $0.025_{0.011}$ & $66.18_{2.41}$ & $1.68\cdot10^7$ \\
         & $15$ & $0.256_{0.083}$ & $0.415_{0.080}$ & $71.30_{0.63}$ & $1.77\cdot10^7$ \\
        \midrule
        \multirow{4}{*}{\makecell[l]{\textit{Simplified} \\ FB: 16 neurons, 1 layer\\ FF: linear}}
         & $6$ & $0.129_{0.016}$ & $0.063_{0.030}$ & $61.20_{1.62}$ & $4.74\cdot10^5$ \\
         & $9$ & $0.137_{0.015}$ & $0.039_{0.009}$ & $68.20_{1.28}$ & $7.05\cdot10^5$ \\
         & $12$ & $0.277_{0.094}$ & $0.132_{0.122}$ & $62.29_{0.73}$ & $9.36\cdot10^5$ \\
         & $15$ & $2.006_{0.048}$ & $1.026_{0.054}$ & $72.69_{1.11}$ & $1.17\cdot10^6$ \\
        \bottomrule
    \end{tabular}
    \caption{Results of the pretraining and PRC experiments performed with the simplified controller architecture for the sMNIST task. The modalities and metrics are those already described in the paper, averaging the results (reported as $\mathrm{mean_{std \, dev}}$) from 4 runs with different random seeds. For each scenario, the digital cost is quantified by the number of FLOPs. A comparison with the ``original'' controllers adopted in our work is also shown.}
    \label{tab:simpler_controller_smnist}
\end{table}

\begin{table}[!ht]
    \centering
    \setlength{\tabcolsep}{6pt}
    \renewcommand{\arraystretch}{1.0}
    \begin{tabular}{llcccc}
        \toprule
        \multirow{2}{*}{\makecell[l]{\textbf{Controller}\\ \textbf{configuration}}} & \multirow{2}{*}{$\bm{n_y}$} & \multicolumn{2}{c}{\textbf{Pretraining}} & \multirow{2}{*}{\makecell{\textbf{PRC performance}\\ \textbf{(NRMSE)}}} & \multirow{2}{*}{\textbf{FLOPs}} \\
        \cmidrule(lr){3-4}
        & & \textbf{Test RMSE} & \textbf{Trajectory NRMSE} & & \\
        \midrule
        \multirow{4}{*}{\makecell[l]{\textit{``Original'' (in our work)} \\ FB: 64 neurons, 2 layers\\ FF: 64 neurons, 2 layers}}
         & $6$  & $0.010_{0.002}$ & $0.048_{0.001}$ & $0.525_{0.003}$ & $3.82\cdot10^7$ \\
         & $9$  & $0.117_{0.005}$ & $0.052_{0.006}$ & $0.505_{0.016}$ & $4.05\cdot10^7$ \\
         & $12$ & $0.192_{0.021}$ & $0.079_{0.003}$ & $0.435_{0.014}$ & $4.28\cdot10^7$ \\
         & $15$ & $0.663_{0.006}$ & $0.187_{0.007}$ & $0.425_{0.006}$ & $4.51\cdot10^7$ \\
        \midrule
        \multirow{4}{*}{\makecell[l]{\textit{Simplified} \\ FB: 16 neurons, 1 layer\\ FF: linear}}
         & $6$ & $0.030_{0.006}$ & $0.050_{0.001}$ & $0.536_{0.008}$ & $1.23\cdot10^6$ \\
         & $9$ & $0.198_{0.003}$ & $0.056_{0.001}$ & $0.509_{0.002}$ & $1.83\cdot10^6$ \\
         & $12$ & $0.445_{0.004}$ & $0.129_{0.001}$ & $0.495_{0.006}$ & $2.43\cdot10^6$ \\
         & $15$ & $0.912_{0.022}$ & $0.244_{0.030}$ & $0.491_{0.011}$ & $3.03\cdot10^6$ \\
        \bottomrule
    \end{tabular}
    \caption{Results of the pretraining and PRC experiments performed with the simplified controller architecture for the Mackey--Glass task. The modalities and metrics are those already described in the paper, averaging the results (reported as $\mathrm{mean_{std \, dev}}$) from 4 runs with different random seeds. For each scenario, the digital cost is quantified by the number of FLOPs. A comparison with the ``original'' controllers adopted in our work is also shown.}
    \label{tab:simpler_controller_MG}
\end{table}

To demonstrate that this simplified controller parametrization still exhibits good PRC performance, we also report quantitative results Tabs.~\ref{tab:simpler_controller_smnist} and \ref{tab:simpler_controller_MG}. We considered only sMNIST (Tab.~\ref{tab:simpler_controller_smnist}) and Mackey--Glass (Tab.~\ref{tab:simpler_controller_MG}) tasks, and, for each one of them, we performed the physical reservoir pretraining stage followed by the readout layer training, using the same modalities adopted in the paper. The evaluation metrics are the same as defined in the main paper, and the tables also report the number of FLOPs for each scenario. A comparison with the ``original'' controllers adopted in our work (2 layers of 64 neurons each) is shown in the tables as well. As can be observed, the number of FLOPs decreases significantly with the new simplified architecture (by one or two orders of magnitude), while the PRC performance does not appear to suffer a substantial degradation, and in some cases even improves over the original architecture. A similar trend can be observed during the pretraining stage, where the original architecture achieves better metrics, although the difference remains relatively limited.

These results suggest that simpler control laws can also be effective within the proposed pretraining framework, improving digital efficiency even for small reservoir dimensions $n_y$. A systematic optimization of the controller architecture is left for future work.

\section{Cross-Task Physical Reservoir Computing Experiments}
\label{app:cross-task_experiments}
The main benchmark experiments use task-specific RON configurations (e.g., spectral radius and time step), as specified in Appendix~\ref{app:dataset_collection} and consistent with established reservoir-tuning practice~\cite{echo_state_network_tutorial,exploration_of_effects_of_different_network_topologies_on_esn,optimization_of_leaky_esn,esn_with_band_pass_neurons,random_oscillators_network_for_time_series_processing}. Consequently, both the digital reference dynamics and the physical reservoirs optimized to reproduce them are specialized to individual tasks. This appendix presents additional experiments assessing whether these physical reservoirs can be reused across tasks. The results show that the optimized reservoirs remain effective on multiple tasks without reoptimizing their morphology, with only modest performance degradation in most cases.

This distinction is important for interpreting the proposed approach. A pretrained physical reservoir that remains effective on unseen tasks supports the hypothesis that co-optimization produces a reusable computational substrate with dynamics broadly useful for physical reservoir computing, rather than merely implementing an indirect form of task-specific training. Cross-task reuse also addresses a practical deployment constraint: in PRC, the reservoir is a physical system whose parameters are fixed after fabrication. If pretraining transferred only to its target task, deployment on each new task would require co-optimizing and fabricating a new substrate.

\paragraph{Experimental setup} We restrict these experiments to reservoirs of dimension $n_y=6$. In each experiment, a \emph{source} physical reservoir---comprising the soft robots, map, and controllers obtained by co-optimization against a reference RON tuned for a specific \emph{source} task---is combined with a readout layer trained on a different \emph{target}-task dataset.

To address cases in which the source and target input sequences $\vect{u}\in\R^{n_u}$ have different dimensions, and the source feedforward (FF) controller therefore cannot be reused, we consider a second scenario in which the FF controller is re-tuned for the target task. Specifically, we fix the soft robots ($\theta_\mathrm{rob}$), map ($\theta_\mathrm{map}$), and feedback controller ($\theta_\mathrm{contr}^{(\mathrm{fb})}$) obtained by co-optimization with the source-task reference RON, and perform a brief auxiliary pretraining stage for only the FF controller ($\theta_\mathrm{contr}^{(\mathrm{ff})}$) using the target-task reference RON.

Finally, we evaluate a third scenario in which the map ($\theta_\mathrm{map}$) and feedback controller ($\theta_\mathrm{contr}^{(\mathrm{fb})}$), in addition to the FF controller, are re-tuned for the target task during auxiliary pretraining. 

None of these experiments modifies the soft robots pretrained for the source task. The same physical substrate can therefore be reused across tasks by reprogramming only its digital components.

We evaluate all 12 cross-task source--target combinations among the four tasks considered in this work: sMNIST, ADIAC, Mackey--Glass, and Lorenz96. The target-task input sequence $\vect{u}\in\R^{n_u}$ is supplied to the source reservoir using the time-step parameter $\Delta t$ of the target-task reservoir. With the source physical reservoir fixed, we conduct the target-task PRC experiments according to the protocol described in Sec.~5 of the paper. Each source--target configuration is evaluated over four runs with different random seeds, and the results are aggregated across runs.

\begin{table}[!b]
    \small
    \centering
    \setlength{\tabcolsep}{6pt}
    \renewcommand{\arraystretch}{1.08}
    \newcommand{\fftunedcell}[1]{\cellcolor{blue!6}#1}
    \newcommand{\alltunedcell}[1]{\cellcolor{blue!12}#1}
    \begin{tabular}{@{}llcccc@{}}
        \toprule
        \multirow{2}{*}{\makecell[l]{\textbf{Source}\\\textbf{reservoir}}}
        & \multirow{2}{*}{\makecell[l]{\textbf{Target-task}\\\textbf{tuning}}}
        & \multicolumn{4}{c}{\textbf{Target task}} \\
        \cmidrule(lr){3-6}
        & & \makecell{sMNIST \\ $[\%]$ ($\uparrow$)}
            & \makecell{ADIAC \\ $[\%]$ ($\uparrow$)}
            & \makecell{Mackey--Glass\\ (NRMSE [-]) ($\downarrow$)}
            & \makecell{Lorenz96\\ (NRMSE [-]) ($\downarrow$)} \\
        \midrule
        \multirow{3}{*}{sMNIST}
            & None & $64.03^\dagger$ & $21.85_{2.34}$ & $0.592_{0.017}$ & -- \\
            & \fftunedcell{FF only} & \fftunedcell{--} & \fftunedcell{$26.31_{0.41}$} & \fftunedcell{$0.619_{0.014}$} & \fftunedcell{$0.768_{0.054}$} \\
            & \alltunedcell{FF + FB + map} & \alltunedcell{--} & \alltunedcell{$44.58_{0.88}$} & \alltunedcell{$0.527_{0.004}$} & \alltunedcell{$0.716_{0.144}$} \\
        \midrule
        \multirow{3}{*}{ADIAC}
            & None & $54.77_{2.35}$ & $44.58^\dagger$ & $0.540_{0.022}$ & -- \\
            & \fftunedcell{FF only} & \fftunedcell{$37.46_{3.88}$} & \fftunedcell{--} & \fftunedcell{$0.584_{0.005}$} & \fftunedcell{$0.765_{0.088}$} \\
            & \alltunedcell{FF + FB + map} & \alltunedcell{$63.14_{0.75}$} & \alltunedcell{--} & \alltunedcell{$0.527_{0.001}$} & \alltunedcell{$0.666_{0.103}$} \\
        \midrule
        \multirow{3}{*}{Mackey--Glass}
            & None & $62.33_{0.58}$ & $25.54_{4.30}$ & $0.525^\dagger$ & -- \\
            & \fftunedcell{FF only} & \fftunedcell{$61.33_{2.02}$} & \fftunedcell{$33.75_{1.75}$} & \fftunedcell{--} & \fftunedcell{$0.753_{0.090}$} \\
            & \alltunedcell{FF + FB + map} & \alltunedcell{$62.52_{2.02}$} & \alltunedcell{$43.10_{1.62}$} & \alltunedcell{--} & \alltunedcell{$0.742_{0.118}$} \\
        \midrule
        \multirow{3}{*}{Lorenz96}
            & None & -- & -- & -- & $0.567^\dagger$ \\
            & \fftunedcell{FF only} & \fftunedcell{$64.84_{2.78}$} & \fftunedcell{$37.62_{1.36}$} & \fftunedcell{$0.535_{0.036}$} & \fftunedcell{--} \\
            & \alltunedcell{FF + FB + map} & \alltunedcell{$61.53_{3.24}$} & \alltunedcell{$44.82_{0.88}$} & \alltunedcell{$0.523_{0.004}$} & \alltunedcell{--} \\
        \bottomrule
    \end{tabular}
    \caption{Results of the cross-task experiments. Each three-row block reports the performance of a source-task reservoir on the four target tasks. The target-task tuning column identifies the digital components re-tuned against the target-task reference RON: ``None'' denotes no auxiliary pretraining, ``FF only'' denotes re-tuning only the feedforward controller, and ``FF + FB + map'' denotes re-tuning both controllers and the map. The light-blue shading distinguishes the three cases also visually Entries show the mean and standard deviation ($\mathrm{mean_{std \, dev}}$) across four random seeds, using classification accuracy for classification tasks and Normalized Root Mean Square Error (NRMSE) for forecasting tasks. A dash denotes a configuration that was not evaluated. Entries marked with $^\dagger$ denote the ``standard'' pretraining experiments in which the source and target tasks coincide.}
    \label{tab:cross-task_experiments}
\end{table}

\paragraph{Results} Tab.~\ref{tab:cross-task_experiments} summarizes the cross-task results. For each source--target pair, PRC performance is reported as the mean and standard deviation ($\mathrm{mean_{std \, dev}}$) across four runs.

Without auxiliary target-task-specific pretraining (rows labeled ``None'' in Tab.~\ref{tab:cross-task_experiments}), several source reservoirs transfer effectively to an unseen target task and achieve performance close to that of the corresponding ``standard'' target-task reservoir. This trend is particularly evident for Mackey--Glass and sMNIST, for which the mean performance degradation across the other source reservoirs is only \SI{7.8}{\percent} and \SI{8.6}{\percent}, respectively. For sMNIST, the Lorenz96 source reservoir performs slightly better than the sMNIST source reservoir. This result may reflect a favorable incidental choice of the Lorenz96 reference RON dynamics, which also appear well suited to sMNIST. Conversely, ADIAC is more challenging for non-target-specific reservoirs, with a mean performance degradation of \SI{46.8}{\percent} relative to its target-specific counterpart. Source--target pairs with different input dimensions $n_u$ cannot be evaluated in this scenario because the source FF controller cannot be reused; these configurations are therefore shown as dashes in the table.

In the second scenario (rows labeled ``FF only'' in Tab.~\ref{tab:cross-task_experiments}), re-optimizing only the FF controller does not consistently improve performance. The mean performance degradation is \SI{14.8}{\percent} for sMNIST, \SI{27.0}{\percent} for ADIAC, and \SI{10.3}{\percent} for Mackey--Glass. This strategy nevertheless enables source--target pairings with different input dimensions. In particular, the sMNIST, ADIAC, and Mackey--Glass source reservoirs can be evaluated on Lorenz96, with an average performance degradation of \SI{34.4}{\percent} relative to the corresponding Lorenz96 reservoir.

Finally, re-tuning the map and both controllers---that is, all digital but none of the physical components---during the auxiliary target-task pretraining stage (rows labeled ``FF + FB + map'' in the table) substantially improves performance. The mean performance degradation drops to \SI{2.6}{\percent} for sMNIST, \SI{0.01}{\percent} for ADIAC, \SI{0.001}{\percent} for Mackey--Glass, and \SI{24.9}{\percent} for Lorenz96. Except for Lorenz96, all non-corresponding source reservoirs then achieve approximately the same performance as the corresponding task-specific reservoir. These results highlight the ability of a pretrained physical reservoir to remain effective on unseen tasks.

One possible explanation for the \SI{24.9}{\percent} performance degradation of Lorenz96 target task could be the choice of the input time step $\Delta t$, which is determined by the target-task reservoir and may be suboptimal for the source-reservoir dynamics. However, adopting the source-reservoir value of $\Delta t$ would impose an input sampling rate based solely on the source dynamics, without accounting for relevant characteristics of the target input, such as its duration and frequency content. Future work could therefore tune the input sampling rate according to both the dynamical properties of the source reservoir (e.g., its resonance frequency) and the temporal characteristics of the target input (e.g., its dominant frequency).

\FinishSupplementDocument

\end{document}